%% file: paper.tex
\documentclass[sigconf, screen]{acmart}
\include{macro}

\AtBeginDocument{%
	\providecommand\BibTeX{{%
			\normalfont B\kern-0.5em{\scshape i\kern-0.25em b}\kern-0.8em\TeX}}}

\copyrightyear{2020}
\acmYear{2020}
\setcopyright{rightsretained}
\acmConference[ESEC/FSE '20]{Proceedings of the 28th ACM Joint European Software Engineering Conference and Symposium on the Foundations of Software Engineering}{November 8--13, 2020}{Virtual Event, USA}
\acmBooktitle{Proceedings of the 28th ACM Joint European Software Engineering Conference and Symposium on the Foundations of Software Engineering (ESEC/FSE '20), November 8--13, 2020, Virtual Event, USA}
\acmDOI{10.1145/3368089.3409704}
\acmISBN{978-1-4503-7043-1/20/11}
\acmSubmissionID{fse20main-p260-p}

\acmPrice{}

\begin{document}

\title{Do the Machine Learning Models on a Crowd Sourced Platform Exhibit Bias? An Empirical Study on Model Fairness}


\author{Sumon Biswas}
\affiliation{
	\institution{Dept. of Computer Science, Iowa State University}
	\city{Ames}
	\state{IA}
	\country{USA}}
\email{sumon@iastate.edu}

\author{Hridesh Rajan}
\affiliation{
	\institution{Dept. of Computer Science, Iowa State University}
	\city{Ames}
	\state{IA}
	\country{USA}}
\email{hridesh@iastate.edu}


\renewcommand{\shortauthors}{Biswas and Rajan}

\input{abstract}

\ccsdesc[500]{Software and its engineering~Software creation and management}
\ccsdesc[500]{Computing methodologies~Machine learning}


\keywords{fairness, machine learning, models}


\maketitle

\input{introduction}
\input{background}
\input{methodology}
\input{unfairness}
\input{mitigation}
\input{impact}
\input{threat}
\input{related}
\input{conclusion}


\begin{acks}
This work was supported in part by US NSF under grants CNS-15-13263, and CCF-19-34884. All opinions are of the authors and do not reflect the view of sponsors.
\end{acks}

\balance
\bibliographystyle{ACM-Reference-Format}
\bibliography{refs}


\end{document}

%% file: macro.tex
\usepackage{xcolor} 
\usepackage{multirow}
\usepackage{booktabs} 
\usepackage{amsmath}
\usepackage{graphicx}
\usepackage[belowskip=-12pt]{caption}
\usepackage{rotating}
\usepackage{tabularx}
\usepackage{subfig}
\usepackage{wrapfig}
\usepackage{listings}
\usepackage{color, colortbl}
\usepackage{balance} 
\usepackage{lscape}
\usepackage{hyperref}
\usepackage{xspace}
\usepackage{framed}
\usepackage{syntax}
\usepackage{soul}
\usepackage{flushend}
   
\usepackage[tikz]{bclogo}
\usepackage{natbib}

\usepackage{hhline}
\usepackage{diagbox}
\usepackage{array, makecell}
\usepackage{microtype}
\definecolor{cellgreen}{RGB}{202,236,193}
\definecolor{cellblue}{RGB}{204, 229, 255}
\definecolor{cellpurple}{RGB}{216,191,216}
\definecolor{cellred}{RGB}{255, 204, 204}
\definecolor{cellorange}{RGB}{255, 229, 204}
\definecolor{cellgold}{RGB}{240,230,140}
\definecolor{cellsteel}{RGB}{230,230,250}

\definecolor{codegreen}{rgb}{0,0.6,0}
\definecolor{codegray}{rgb}{0.5,0.5,0.5}
\definecolor{codepurple}{rgb}{0.58,0,0.82}
\definecolor{backcolour}{rgb}{0.95,0.95,0.92}
\definecolor{Gray}{gray}{0.1}

\lstdefinestyle{mystyle}{
	backgroundcolor=\color{backcolour},   
	commentstyle=\color{codegreen},
	keywordstyle=\color{magenta},
	numberstyle=\tiny\color{codegray},
	stringstyle=\color{codepurple},
	basicstyle=\scriptsize,
	breakatwhitespace=false,         
	breaklines=true,                 
	captionpos=b,                    
	keepspaces=true,                 
	numbers=left,                    
	numbersep=5pt,                  
	showspaces=false,                
	showstringspaces=false,
	showtabs=false,                  
	tabsize=2,
	xleftmargin=.013\textwidth
}

\lstdefinelanguage{Pythonna}{%
	language     = python,
	morekeywords = {to_categorical, flow_from_directory, pad_sequences, load_image}
}

\lstdefinestyle{customc}{
	belowcaptionskip=1\baselineskip,
	breaklines=false,
	frame= single,
	breaklines = true,
	xleftmargin=\parindent,
	language= Pythonna,
	showstringspaces=false,
	basicstyle=\footnotesize\ttfamily,
	keywordstyle=\bfseries\color{green!40!black},
	commentstyle=\itshape\color{purple!40!black},
	identifierstyle=\color{blue},
	stringstyle=\color{codegreen},
	backgroundcolor=\color{gray!4}
}

\graphicspath{{./figures/}}

\newcommand{\tabref}[1]{Table~\ref{#1}}
\newcommand{\fignref}[1]{Figure~\ref{#1}}

\newcommand{\secref}[1]{\S\ref{#1}}
\newcommand{\secnref}[1]{\S\ref{#1}}

\newcommand{\etal}{{\em et al.}\xspace}

\newcounter{rqs}
\stepcounter{rqs}

\newcounter{NumObservations}
\stepcounter{NumObservations}
\definecolor{shadecolor}{rgb}{.9,.9,.9}

\newcommand{\finding}[1]{
	\begin{bclogo}[couleur= gray!7, epBord= 0.6, arrondi=0.1, logo=\bcinfo, marge= 2, ombre=false, couleurBord=black!60, tailleOndu=3, sousTitre ={\em #1}]{Finding \arabic{NumObservations}: } 
		
	\end{bclogo}
	\stepcounter{NumObservations}
}

%% file: abstract.tex
\begin{abstract}

Machine learning models are increasingly being used in important decision-making software such as approving bank loans, recommending criminal sentencing, hiring employees, and so on. It is important to ensure the fairness of these models so that no discrimination is made based on \textit{protected attribute} (e.g., race, sex, age) while decision making. Algorithms have been developed to measure unfairness and mitigate them to a certain extent. In this paper, we have focused on the empirical evaluation of fairness and mitigations on real-world machine learning models. We have created a benchmark of 40 top-rated models from Kaggle used for 5 different tasks, and then using a comprehensive set of fairness metrics, evaluated their fairness. Then, we have applied 7 mitigation techniques on these models and analyzed the fairness, mitigation results, and impacts on performance. We have found that some model optimization techniques result in inducing unfairness in the models. On the other hand, although there are some fairness control mechanisms in machine learning libraries, they are not documented. The mitigation algorithm also exhibit common patterns such as mitigation in the post-processing is often costly (in terms of performance) and mitigation in the pre-processing stage is preferred in most cases. We have also presented different trade-off choices of fairness mitigation decisions. Our study suggests future research directions to reduce the gap between theoretical fairness aware algorithms and the software engineering methods to leverage them in practice.

\end{abstract}

%% file: introduction.tex
\section{Introduction}
\label{sec:introduction}

Since machine learning (ML) models are increasingly being used in making important decisions that affect human lives, it is important to ensure that the prediction is not biased toward any protected attribute such as race, sex, age, marital status, etc. 
ML fairness has been studied for about past 10 years \cite{friedler2019comparative}, and several fairness metrics and mitigation techniques \cite{kamiran2012data, feldman2015certifying, zafar2015fairness, calders2010three, chouldechova2017fair, zhang2018mitigating, kamishima2012fairness, hardt2016equality, zhang2018mitigating} have been proposed. Many testing strategies have been developed \cite{udeshi2018automated, aggarwal2019black, galhotra2017fairness} to detect unfairness in software systems. Recently, a few tools have been proposed \cite{adebayo2016fairml, bellamy2018ai, tramer2017fairtest, sokol2019fat} to enhance fairness of ML classifiers. 
However, we are not aware how much fairness issues exist in ML models from practice. Do the models exhibit bias? If yes, what are the different bias types and what are the model constructs related to the bias? Also, is there a pattern of fairness measures when different mitigation algorithms are applied? In this paper, we have conducted an empirical study on ML models to understand these characteristics. 

Harrison \etal studied how ML model fairness is perceived by 502 Mechanical Turk workers \cite{harrison2020empirical}. 
Recently, Holstein \etal conducted an empirical study on ML fairness by surveying and interviewing industry practitioners \cite{holstein2019improving}. They outlined the challenges faced by the developers and the support they need to build fair ML systems. They also discussed that it is important to understand the fairness of existing ML models and improve software engineering to achieve fairness. In this paper, we have analyzed the fairness of 40 ML models collected from a crowd sourced platform, Kaggle, and answered the following research questions. 

\noindent\textbf{RQ1: (Unfairness)} What are the unfairness measures of the ML models in the wild, and which of them are more or less prone to bias?

\noindent\textbf{RQ2: (Bias mitigation)} What are the root causes of the bias in ML models, and what kind of techniques can successfully mitigate those bias?

\noindent\textbf{RQ3: (Impact)} What are the impacts of applying different bias mitigating techniques on ML models? 

First, we have created a benchmark of ML models collected from Kaggle. We have manually verified the models and selected appropriate ones for the analysis. Second, we have designed an experimental setup to measure, achieve, and report fairness of the ML models. Then we have analyzed the result to answer the research questions. 
The key findings are: 
model optimization goals are configured towards overall performance improvement, causing unfairness. A few model constructs are directly related to fairness of the model. However, ML libraries do not explicitly mention fairness in documentation.
Models with effective pre-processing mitigation algorithm are more reliable and pre-processing mitigations always retain performance. 
We have also reported different patterns of exhibiting bias and mitigating them.
Finally, we have reported the trade-off concerns evident for  those models.

The paper is organized as follows: \secnref{sec:background} describes the background and necessary terminology used in this paper. In \secref{sec:methodology}, we have described the methodology of creating the benchmark and setting up experiment, and discussed the fairness metrics and mitigation techniques. \secref{sec:unfairness} describes the fairness comparison of the models, \secref{sec:mitigation} describes the mitigation techniques, and \secref{impact} describes the impacts of mitigation. We have discussed the threats to validity in \secref{sec:threats}, described the related work in \secref{sec:related}, and concluded in \secref{sec:conc}.

%% file: background.tex
\section{Background}
\label{sec:background}
The basic idea of ML fairness is that the model should not discriminate between different individuals or groups from the protected attribute class \cite{friedler2019comparative, galhotra2017fairness}. \textit{Protected attribute} (e.g., race, sex, age, religion) is an input feature, which should not affect the decision making of the models solely. Chen \etal listed 12 protected attributes for fairness analysis~\cite{chen2019fairness}.
One trivial idea is to remove the protected attribute from the dataset and use that as training data. Pedreshi \etal showed that due to the redundant encoding of training data, it is possible that protected attribute is propagated to other correlated attributes \cite{pedreshi2008discrimination}. Therefore, we need fairness aware algorithms to avoid bias in ML models.
In this paper, we have considered both group fairness and individual fairness. \textit{Group fairness} measures whether the model prediction discriminates between different groups in the protected attribute class (e.g., sex: \textit{male/female}) \cite{dwork2012fairness}. \textit{Individual fairness} measures whether similar prediction is made for similar individuals those are only different in protected attribute \cite{dwork2012fairness}. Based on different definitions of fairness, many group and individual fairness metrics have been proposed. Additionally, many fairness mitigation techniques have been developed to remove unfairness or bias from the model prediction. The fairness metrics and mitigation techniques have been described in the next section.

%% file: methodology.tex
\section{Methodology}
\label{sec:methodology}
In this section, first, we have described the methodology to create the benchmark of ML models for fairness analysis. Then we have described our experiment design and setup. Finally, we have discussed the fairness metrics we evaluated and mitigation algorithms we applied on each model.

\subsection{Benchmark Collection}
\label{subsec:benchmark}

\begin{figure}[t]
	\centering
	\includegraphics[width=\columnwidth]{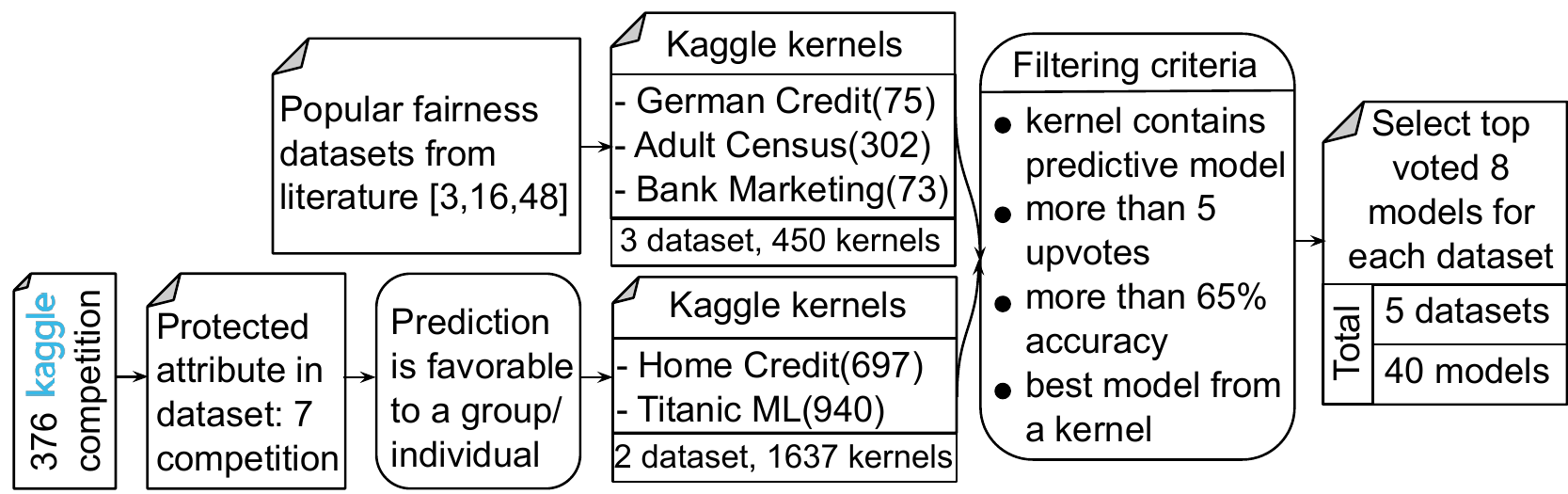}
	\caption{Benchmark model collection process}
	\label{benchmark}
\end{figure}

We have collected ML models from Kaggle kernels \cite{kaggle}.
Kaggle is one of the most popular data science (DS) platform owned by Google. 
Data scientists, researchers, and developers can host or take part in DS 
competition, share dataset, task, and solution. 
Many Kaggle solutions resulted in impactful ML algorithms and research 
such as neural networks used by Geoffrey Hinton and George Dahl \cite{dahl2014multi}, 
improving the search for the Higgs Boson at CERN \cite{higs-boson}, 
state-of-the-art HIV research \cite{carpenter2011may}, etc. 
There are 376 competitions and 28,622 datasets in Kaggle to date. 
The users can submit solutions for the competitions and dataset-specific tasks. 
To create a benchmark to analyze the fairness of ML models, we have collected 
40 kernels from the Kaggle. 
Each kernel provides solution (code and description) for a specific data science task.
In this study, we have analyzed ML models that operate on 1) datasets utilized by prior studies on fairness, and 2) datasets with protected attribute (e.g., sex, race). With this goal, we have collected the ML models with different filtering criteria for each category. The overall process of collecting the benchmark has been depicted in \fignref{benchmark}.

To identify the datasets used in prior fairness studies, we refer to the work on fairness testing by Galhotra \etal \cite{galhotra2017fairness}, where two datasets, German Credit and Adult Census have been used. Udeshi \etal experimented on models for the Adult Census dataset \cite{udeshi2018automated}. 
Aggarwal \etal used six datasets: German Credit, Adult Census, Bank Marketing, US Executions, Fraud Detection, and Raw Car Rentals) \cite{aggarwal2019black}. 
Among these datasets, German Credit, Adult Census and Bank Marketing dataset are available on Kaggle. From the solutions for these datasets, we have collected 440 kernels (65 for German Credit, 
302 for Adult Census, and 
73 for Bank Marketing).
Furthermore, we have filtered the kernels based on three criteria to select the top-rated ones: 1) contain predictive models (some kernels only contain exploratory data analysis), 2) at least 5 upvotes, and 3) accuracy $\ge$ 65\%. 
Often a kernel contains multiple models and tries to find the best performing one. 
In these cases, we have selected the best performing model from every kernel. Thus, we have selected the top 8 models based on upvotes for each of the 3 datasets and got 24 ML models.

Chen \etal \cite{chen2019fairness} listed 12 protected attributes, e.g., age, sex, race, etc. for fairness analysis. We have found 7 competitions in Kaggle, that contain any of these attributes.
From the selected ones, we have filtered out the competitions that involve prediction decisions not being favorable to individuals or a specific group. 
For example, although this competition \cite{cspr} has customers \textit{age} and \textit{sex} in the dataset, the classification task is to recommend an appropriate product to the customers, which we can not classify as fair or unfair.
Thus, we have got two appropriate competitions 
with several kernels. To select ML models from these competitions, we have utilized the same filtering criteria used before and selected 8 models for each dataset based on the upvotes.
Finally, we have created a 
benchmark containing 40 top-rated Kaggle models that operate on 5 datasets.
The characteristics of the datasets and tasks in the benchmark are shown in \tabref{main-table}.

\begin{table*}[]
  \vspace{1.3mm}
  \centering
  \setlength\tabcolsep{5pt} 
  \caption{The datasets used in the fairness experimentation. \# F: Feature count. PA: Protected attribute.}
    \begin{tabular}{|p{2.6cm}|r|r|p{.5cm}|p{11.4cm}|}
    \hline
    \textbf{Dataset} & \multicolumn{1}{c|}{\textbf{Size}} & \multicolumn{1}{c|}{\textbf{\# F}} & \multicolumn{1}{c|}{\textbf{PA}} & \textbf{Description} \\
    \hline
    German Credit~\cite{gc} & 1,000  & 21    & age, sex & {This dataset contains personal information about individuals and predicts credit risk (good or bad credit). The \textit{age} protected attribute is categorized into young ($< 25$) and old ($\geq 25$) based on \cite{friedler2019comparative}.}\\
    \hline
    Adult Census~\cite{adult} & 32,561 & 12    & race, sex &{This dataset comprises of individual information from the 1994 U.S. census. The target feature of this dataset is to predict whether an individual earns $\ge\$50,000$ or not in a year. }\\
    \hline
    Bank Marketing~\cite{bank} & 41,188 & 20    & age   & {This dataset contains the direct marketing campaigns data of a Portuguese bank. The goal is to predict whether a client will subscribe for a term deposit or not.}\\
    \hline
    Home Credit~\cite{home} & 3,075,11 & 240    & sex   & {This dataset contains data related to loan applications for individuals who do not get loan from the traditional banks. The target feature is to predict whether an individual who can repay the loan, get the application accepted or not.}\\
    \hline
    Titanic ML \cite{titanic} & 891 & 10    & sex   & {This dataset contains data about the passengers of Titanic. The target feature is to predict whether the passenger survived the sinking of Titanic or not. The target of the test set is not published. So, we have taken the training data and further split it into train and test.} \\
    \hline
    \end{tabular}%
  \label{main-table}%
\end{table*}%

\begin{figure*}[t]
	\centering
	\includegraphics[width=.9\linewidth]{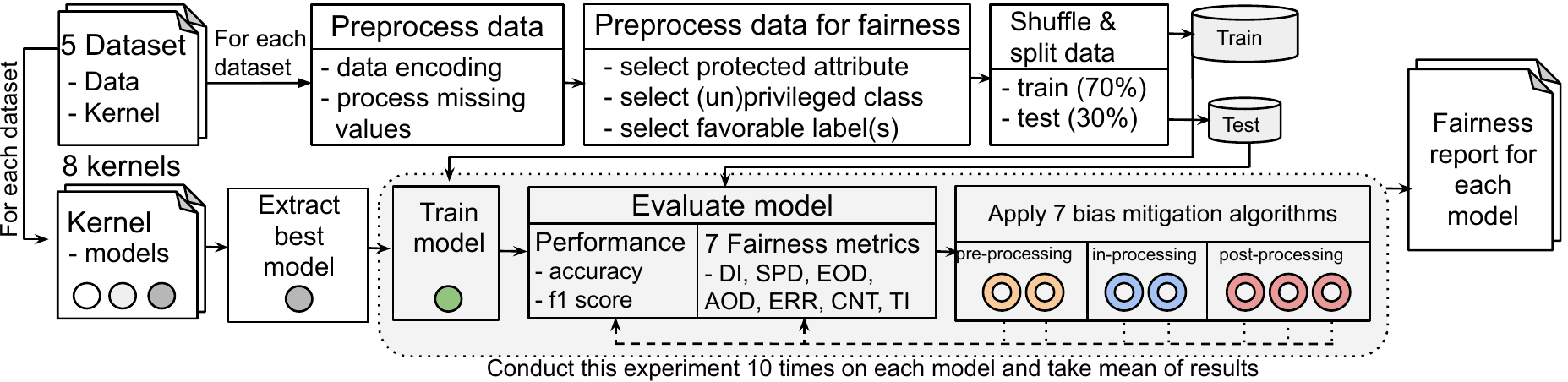}
	\caption{Experimentation to compute performance, fairness and mitigation impacts of machine learning models.}
	\label{experiment}
\end{figure*}

\subsection{Experiment Design}
\label{subsec:experiemnt}

After creating the benchmark, we have experimented on the models, 
evaluated performance and fairness metrics, and applied different bias mitigation techniques to observe the impacts. Our experiment design process is shown in \fignref{experiment}. The experiments on the benchmark have been peer reviewed and published as an artifact \cite{biswas20203912064}.

In our benchmark, we have models from five dataset categories. 
To be able to compare the fairness of different models in each dataset 
category, we have used the same data preprocessing strategy. 
We have processed the missing or invalid values, transformed continuous 
features to categorical (e.g., age$<$25: young, age$\geq$25: old), and 
converted non-numerical features to numerical (e.g., \textit{female}: 0, \textit{male}: 1). 
We have done some further preprocessing to the dataset to be used for fairness 
analysis: specify the protected attributes, privileged and unprivileged group, 
and what are the favorable label or outcome of the prediction. 
For example, in the Home Credit dataset, \textit{sex} is the protected 
attribute, where \textit{male} is the privileged group, \textit{female} is the unprivileged 
group, and the prediction label is credit risk of the person i.e., good 
(favorable label) or bad. 
For all the datasets, we have used shuffling and same train-test splitting (70\%-30\%) before feeding the data to the models.

For each dataset category, we have eight Kaggle kernels. 
The kernels contain solution code written in Python for solving 
classification problems. 
In general, the kernels follow these stages: data exploration, 
preprocessing, feature selection, modeling, training, evaluation, 
and prediction. 
From the kernels, we have manually extracted the code for modeling, 
training, and evaluation. 
For example, this kernel \cite{kernel-gcra} 
loads the German Credit dataset, performs exploratory analysis and 
selects a subset of the features for training, preprocesses data, and finally 
implements XGBoost classifier for predicting the credit 
risk of individuals. 
We have manually sliced the code for modeling, training, and evaluation. 
Often the kernels try multiple models, evaluate results, and find the best model. 
From a single kernel, we have only sliced the best performing model found by the kernel. 
Some kernels do not specify the best model. 
In this case, we have selected the model with the best accuracy. 
For example, this kernel \cite{kernel-mmta}
works on Adult Census dataset and implements four models 
(Logistic Regression, Decision Tree, K-Nearest Neighbor and Gradient 
Boosting) for predicting income of individuals. 
We have selected the Gradient Boosting classifier model since it gives 
the best accuracy.

After extracting the best model, we train the model and evaluate 
performance (accuracy, F1 score). 
We have found that the model performance in our experiment is 
consistent with the prediction made in the kernel. 
Then, we have evaluated 7 different fairness metrics described in \secref{subsub:fairness-measure}.
Next, we have applied 7 different bias mitigation algorithms separately 
and evaluated the performance and fairness metrics. 
Thus, we collect the result of 9 metrics (2 performance metric, 7 fairness 
metric) before applying any mitigation algorithm and after applying 
each mitigation algorithm. 
For each model, we have done this experiment 10 times and taken the mean of the results as suggested by \cite{friedler2019comparative}. 
We have used the open-source Python library AIF 360 ~\cite{bellamy2018ai} 
developed by IBM for fairness metrics and bias mitigation algorithms. 
All experiments have been executed on a MAC OS 10.15.2, having 4.2 GHz 
Intel Core i7 processor with 32 GB RAM and Python 3.7.6. 

\subsection{Measures}
\label{subsec:measures}

We have computed the algorithmic fairness of each subject model in our benchmark.
Let, $D=(X, Y, Z)$ be a dataset where $X$ is the training data, $Y$ is the binary classification label ($Y=1$ if the label is favorable, otherwise $Y=0$), $Z$ is the protected attribute ($Z = 1$ for privileged group, otherwise $Z = 0$), and $\hat{Y}$ is the prediction label (1 for favorable decision and 0 for unfavorable decision). If there are multiple groups for protected attributes, we have employed a binary grouping strategy (e.g., race attribute in Adult Census dataset has been changed to white/non-white).

\subsubsection{Accuracy Measure}
\label{subsub:accuracy-measure}

Before measuring the fairness of the model, we have computed the performance in terms of accuracy, and F1 score.\\
\textit{Accuracy}: Accuracy is given by the ratio of truly classified items and total number of items.
	\[\text{Accuracy} = (\# \text{ True positive} + \# \text{ True negative}) / \# \text{ Total} \]
\textit{F1 Score}: This metric is given by the harmonic mean of precision and recall.
	\[ \text{F1} = 2 * (\text{Precision} * \text{Recall}) / (\text{Precision} + \text{Recall}) \]
	
\subsubsection{Fairness Measure}
\label{subsub:fairness-measure}

Many quantitative fairness metrics have been proposed in the literature \cite{binns2017fairness} based on different definitions of fairness. For example, AIF 360 toolkit has APIs for computing 71 fairness metrics \cite{bellamy2018ai}. In this paper, without being exhaustive, a representative list of metrics have been selected to evaluate the fairness of ML models. We have adopted the metrics recommendation of Friedler \etal \cite{friedler2019comparative} and further added the individual fairness metrics. \\
\\
\textbf{Metrics based on base rates:} \\ 
\textit{Disparate Impact (DI):} This metric is given by the ratio between the probability of unprivileged group gets favorable prediction and the probability of privileged group gets favorable prediction \cite{feldman2015certifying, zafar2015fairness}. 
	\[ \mathrm{DI} = \mathsf{P} [\hat{Y} = 1 | Z = 0] / \mathsf{P}[\hat{Y} = 1 | Z = 1] \]

\noindent \textit{Statistical Parity Difference (SPD):} This measure is similar to DI but instead of the ratio of probabilities, difference is calculated \cite{calders2010three}.
	\[ \mathrm{SPD} = \mathsf{P}[\hat{Y} = 1 | Z = 0] - \mathsf{P}[\hat{Y} =1 | Z = 1] \]
\textbf{Metrics based on group conditioned rates:} \\
\textit{Equal Opportunity Difference (EOD):} This is given by the true-positive rate (TPR) difference between unprivileged and privileged groups.
\begin{gather*}  
	\mathrm{TPR}_u = \mathsf{P}[\hat{Y} = 1 | Y = 1, Z = 0] \text{ ; } 
	\mathrm{TPR}_p = \mathsf{P}[\hat{Y} = 1 | Y = 1, Z = 1] \\
	\mathrm{EOD} = TPR_u - TPR_p
\end{gather*}

\noindent \textit{Average Odds Difference (AOD):} This is given by the average of false-positive rate (FPR) difference and true-positive rate difference between unprivileged and privileged groups \cite{hardt2016equality}.
	\begin{gather*}  
	\mathrm{FPR}_u = \mathsf{P}[\hat{Y} = 1 | Y = 0, Z = 0] \text{ ;  }
	\mathrm{FPR}_p = \mathsf{P}[\hat{Y} = 1 | Y = 0, Z = 1] \\
	\mathrm{AOD} = \frac{1}{2} \{ (FPR_u - FPR_p) + (TPR_u - TPR_p) \}
	\end{gather*} 
	
\noindent  \textit{Error Rate Difference (ERD):} Error rate is given by the addition of false-positive rate (FPR) and false-negative rate (FNR) \cite{chouldechova2017fair}.  
\begin{align*} 
\mathrm{ERR} &= \text{FPR} + \text{FNR} \\
\mathrm{ERD} &= \text{ERR}_u - \text{ERR}_p 
\end{align*}

\noindent \textbf{Metrics based on individual fairness:} \\
\textit{Consistency (CNT):} This individual fairness metric measures how similar the predictions are when the instances are similar \cite{zemel2013learning}. Here, $n\_neighbors$ is the number of neighbors for the KNN algorithm.
	\[ \mathrm{CNT} =  1 - \frac{1}{n * n\_neighbors} \sum_{i=1}^{n} |\hat{y}_i - \sum_{j\in \mathcal{N}_{n\_neighbors(x_i)}} \hat{y}_j | \]

\textit{Theil Index (TI):} This metric is also called the entropy index which measures both the group and individual fairness \cite{speicher2018unified}. Theil index is given by the following equation where $b_i=\hat{y_i}-y_i+1$.
	\[ \mathrm{TI} = \frac{1}{n} \sum_{i=1}^{n} \frac{b_i}{\mu}\text{ln}\frac{b_i}{\mu} \]

\subsection{Bias Mitigation Techniques} 
\label{subsec:bias-mitigation}
In this section, we have discussed the bias mitigation techniques that have been applied to the models. These techniques can be broadly classified into preprocessing, in-processing, and postprocessing approaches.

\paragraph{Preprocessing Algorithms}
Preprocessing algorithms do not change the model and only work on the dataset before training so that models can produce fairer predictions.

\noindent Reweighing ~\cite{kamiran2012data}: In a biased dataset, different weights are assigned to reduce the effect of favoritism of a specific group. If a class of input has been favored, then a lower weight is assigned in comparison to the class not been favored.

\noindent Disparate Impact Remover ~\cite{feldman2015certifying}: This algorithm is based on the concept of the metric DI that measures the fraction of individuals achieves positive outcomes from an unprivileged group in comparison to the privileged group. To remove the bias, this technique modifies the value of protected attribute to remove distinguishing factors. 

\paragraph{In-processing Algorithms}
In-processing algorithms modify the ML model to mitigate the bias in the original model prediction.

\noindent Adversarial Debiasing ~\cite{zhang2018mitigating}: This approach modifies the ML model by introducing backward feedback (negative gradient) for predicting the protected attribute. This is achieved by incorporating an adversarial model that learns the difference between protected and other attributes that can be utilized to mitigate the bias.

\noindent Prejudice Remover Regularizer ~\cite{kamishima2012fairness}: If an ML model relies on the decision based on the protected attribute, we call that direct prejudice. In order to remove that, one could simply remove the protected attribute or regulate the effect in the ML model. This technique applies the latter approach, where a regularizer is implemented that computes the effect of the protected attribute.

\paragraph{Post-processing Algorithms} 
This genre of techniques modifies the prediction result instead of the ML models or the input data.

\noindent Equalized Odds (E)~\cite{hardt2016equality}: This approach changes the output labels to optimize the EOD metric. In this approach, a linear program is solved to obtain the probabilities of modifying prediction.

\noindent Calibrated Equalized Odds ~\cite{pleiss2017fairness}: To achieve fairness, this technique also optimizes EOD metric by using the calibrated prediction score produced by the classifier.

\noindent Reject Option Classification ~\cite{kamiran2012decision}: This technique favors the instances in privileged group over unprivileged ones that lie in the decision boundary with high uncertainty.

%% file: unfairness.tex
\section{Unfairness in ML Models}
\label{sec:unfairness}

\begin{figure*}[!ht]
	\centering
	\includegraphics[width=\linewidth]{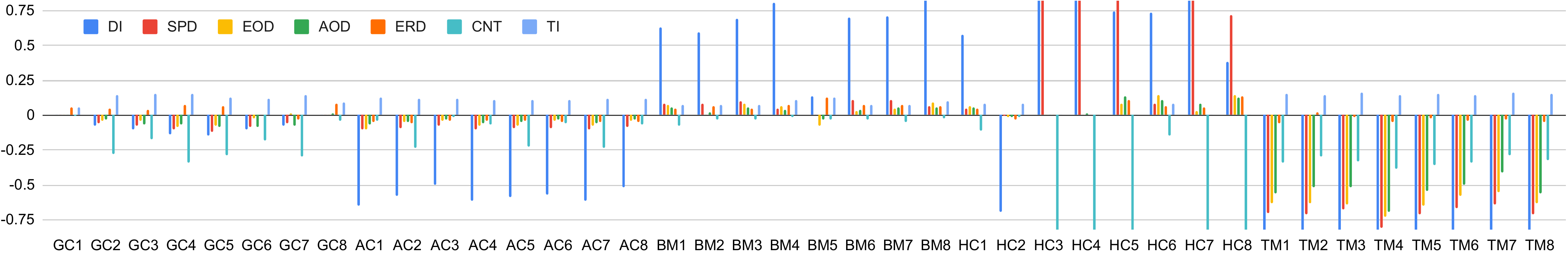}
	\caption{Unfairness exhibited by the ML models with respect to different metrics}
	\label{unfairness}
\end{figure*}

In this section, we have explored the answer of RQ1 by analyzing different fairness measures exhibited by the ML models in our benchmark. 
Do the models have bias in their prediction? If so, which models are fairer and which are more biased? What is causing the models to be more prone to bias? What kind of fairness metric is sensitive to different models?
To answer these questions, we have conducted experiment on the ML models and computed the fairness metrics. The result is presented in \tabref{tab:main}. 
The unfairness measures for all the 40 models are depicted in \fignref{unfairness}. To be able to compare all the metrics in the same chart, disparate impact (DI), and consistency (CNT) have been plotted in the log scale. If the value of a fairness metric is $0$, there is no bias in the model according to the corresponding metric. If the measure is less than or greater than 0, bias exists. The negative bias denotes that the prediction is biased towards privileged group and positive bias denotes that prediction is biased towards unprivileged group.

\begin{figure}[h]
	\centering
	\includegraphics[width=\columnwidth]{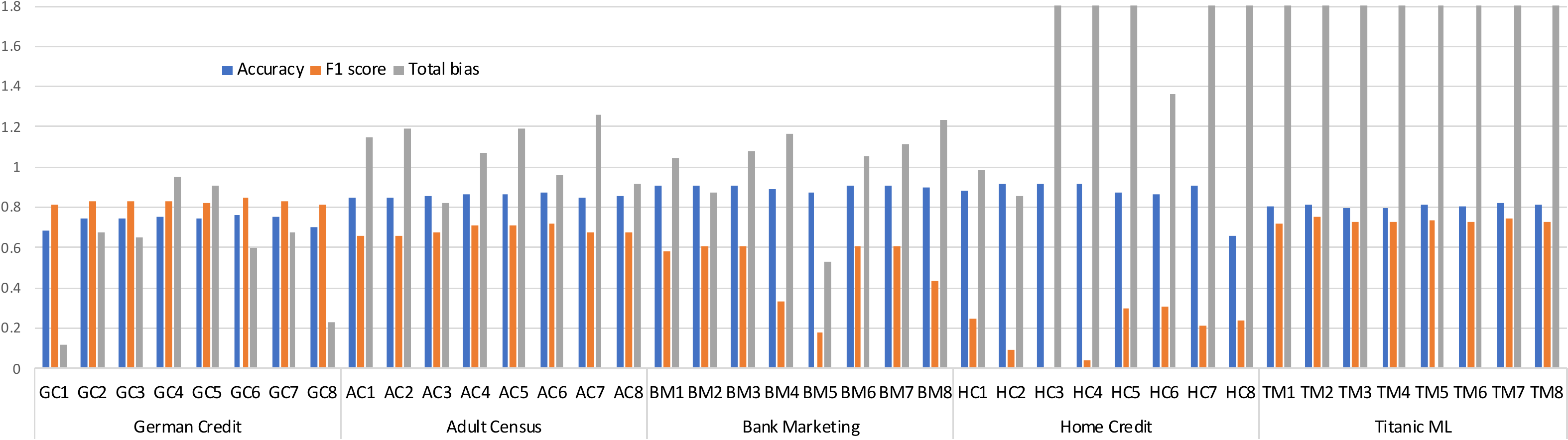}
	\caption{Cumulative bias and performance of the models}
	\label{cumulative}
\end{figure}

We have found that all the models exhibit unfairness and models specific to a dataset show similar bias patterns.
From \fignref{unfairness}, we can see that all the models exhibit bias with respect to most of the fairness metrics. For a model, metric values vary since the metrics follow different definitions of fairness. Therefore, we have compared bias of different models both cumulatively and using the specific metric individually. To compare total bias across all the metrics, we have taken the absolute value of the measures and computed the sum of bias for each model. In \fignref{cumulative}, we can see the total bias exhibited by the models. Although the bias exhibited by models for each dataset follow similar pattern, certain models are fairer than others. \\

\finding{Model optimization goals seek overall performance improvement, which is causing unfairness.}
Model GC1 exhibits the lowest bias among German Credit models. GC1 is a Random Forest (RFT) classifier model, which is built by using a grid search over a given range of hyperparameters. After the grid search, the best found classifier is:
\begin{lstlisting}[language=Python]
RandomForestClassifier(bootstrap=True, ccp_alpha=0.0, class_weight=None, criterion='gini', max_depth=3, max_features=4, max_leaf_nodes=None, max_samples=None, min_impurity_decrease=0.0, min_impurity_split=None, min_samples_leaf=1, min_samples_split=2, min_weight_fraction_leaf=0.0, n_estimators=25, n_jobs=None, oob_score=False, random_state=2, warm_start=False)
\end{lstlisting}
We have found that GC6 is also a Random Forest classifier built through grid search. However, GC6 is less fair in terms of cumulative bias (\fignref{cumulative}), and individual metrics (\fignref{unfairness}) except error rate difference (ERD). We have investigated the reason of the fairness differences in these two models by running both of them by changing one hyperparameter at a time.
We have found that the fairness difference is caused by the scoring mechanism used by the two models. GC1 uses \texttt{scoring=\textquotesingle recall\textquotesingle}, whereas GC6 uses \texttt{scoring=\textquotesingle precision\textquotesingle}, as shown in the following code snippet. 
\begin{lstlisting}[language=Python]
# Model GC1
param_grid = {"max_depth": [3,5, 7, 10,None], "n_estimators":[3,5,10,25,50,150], "max_features": [4,7,15,20]}
GC1 = RandomForestClassifier(random_state=2)
grid_search = GridSearchCV(GC1, param_grid=param_grid, cv=5, scoring='recall', verbose=4)
# Model GC6 
params = {'n_estimators':[25,50,100,150,200,500],'max_depth':[0.5,1,5,10],'random_state':[1,10,20,42], 'n_jobs':[1,2]}
GC6 = RandomForestClassifier()
grid_search_cv = GridSearchCV(GC6, params,scoring='precision')
\end{lstlisting}
Further investigation shows, in German Credit dataset, the data rows are personal information about individuals and task is to predict their credit risk. The data items are not balanced when \textit{sex} of the individuals is concerned. The dataset contains 69\% data instances of \textit{male} and 31\% \textit{female} individuals.
When the model is optimized towards recall (GC1) rather than precision (GC6), the total number of true-positives decreases and false-negative increases. Since the number of instances for privileged group (\textit{male}) is more than the unprivileged group (\textit{female}), decrease in the total number of true-positives also  increases the probability of unprivileged group to be classified as favorable. Therefore, the fairness of GC1 is more than GC2, although the accuracy is less. 
Unlike other group fairness metrics, error rate difference (ERD) accounts for false-positive and false-negative rate difference between privileged and unprivileged group. As described before, optimizing the model for recall increases the total number of false-negatives. We have found that the percentage of \textit{male} categorized as favorable is less than the percentage of \textit{female} categorized as favorable.
Therefore, an increase in the overall false-negative also increased the error rate of unprivileged group, which in turn caused GC1 to be more biased than GC2 in terms of ERD.

From the above discussion, we have observed that the model optimization hyperparameter only considers the overall rates of the performance. However, if we split the data instances based on protected attribute groups, then we see the change of rates vary for different groups, which induces bias. The libraries for model construction also do not provide any option to specify model optimization goals specific to protected attributes and make fairer prediction.

Here, we have seen that GC1 has less bias than GC6 by compromising little accuracy. Do all the models achieve fairness by compromising with performance? We have found that models can achieve fairness along with high performance. To compare model performance with the amount of bias, we have plotted the accuracy and F1 score of the models with the cumulative bias in \fignref{cumulative}. 
We can see that GC6 is the most efficient model in terms of performance and has less bias than 5 out of 7 other models in German Credit data. AC6 has more accuracy and F1 score than any other models in Adult Census, and exhibits less bias than AC1, AC2, AC4, AC5, and AC7. Therefore, models can have better performance and fairness at the same time.\\

\finding{Libraries for model creation do not explicitly mention fairness concerns in model constructs.}
From \fignref{unfairness}, we can see that HC1 and HC2 show difference in most of the fairness metrics, while operating on the same dataset i.e., Home Credit. HC2 is fairer than HC1 with respect to all the metrics except DI. From \tabref{tab:main}, we can see that HC1 has positive bias, whereas HC2 exhibit negative bias. This indicates that HC1 is biased towards unprivileged group and HC2 is biased towards privileged group. 
We have found that HC1 and HC2 both are using Light Gradient Boost (LGB) model for prediction. The code for building the two models are:

\begin{lstlisting}[language=Python]
# Model HC1
HC1 = lgb.LGBMClassifier(n_estimators=10000, objective='binary', class_weight='balanced', learning_rate=0.05, reg_alpha=0.1, reg_lambda=0.1, subsample=0.8, n_jobs=-1, random_state=50)
HC1.fit(X_train, y_train, eval_metric = 'auc', categorical_feature = cat_indices, verbose = 200)
# Model HC2
HC2 = LGBMClassifier(n_estimators=4000, learning_rate=0.03, num_leaves=30, colsample_bytree=.8, subsample=.9, max_depth=7, reg_alpha=.1, reg_lambda=.1, min_split_gain=.01, min_child_weight=2, silent=-1, verbose=-1)
HC2.fit(X_train, y_train, eval_metric= 'auc', verbose= 100)
\end{lstlisting}

We have executed both the models with varied hyperparameter combinations and found that \texttt{class\_weight=\textquotesingle balanced\textquotesingle} is causing HC1 not to be biased towards privileged group. By specifying \texttt{class\_weight}, we can set more weight to the data items belonging to an infrequent class. Higher class weight implies that the data items are getting more emphasis in prediction. When the class weight is set to \texttt{balanced}, the model automatically accounts for class imbalance and adjust the weight of data items inversely proportional to the frequency of the class \cite{joshi2017artificial, lgbm}.
In this case, HC1 mitigates the \textit{male-female} imbalance in its prediction. Therefore, it does not exhibit bias towards the privileged group (\textit{male}). On the other hand, HC2 has less bias but it is biased towards privileged group. Although we want models to be fair with respect to all groups and individuals, trade-off might be needed and in some cases, bias toward unprivileged may be a desirable trait. 

We have observed that \texttt{class\_weight} hyperparameter in LGBMClassifier allows developers to control group fairness directly. However, the library documentation of LGB classifier suggests that this parameter is used for improving performance of the models \cite{lgbm, class-weight}.
Though the library documentation mentions about probability calibration of classes to boost the prediction performance using this parameter, however, there is no suggestion regarding the effect on the bias introduced due to the wrong choice of this parameter. 

From the discussions, we can conclude that library developers still do not provide explicit ways to control fairness of the models. Although some parameters directly control the fairness of the models, libraries do not explicitly mention that.

\finding{Standardizing features before training models can help to remove disparity between groups in the protected class.}
From \fignref{unfairness} and \fignref{cumulative}, we observe that except BM5, other models in Bank Marketing exhibit similar unfairness. BM5 is a Support Vector Classifier (SVC) tuned using a grid search over given range of parameters. In the modeling pipeline, before training the best found SVC, the features are transformed using \texttt{StandardScalar}. Below is the model construction code for BM5 with the best found hyperparameters:

\begin{lstlisting}[language=Python]
tuned_parameters = [{'kernel': ['rbf'], 'gamma': [0.1], 'C': [1]}]
SVC = GridSearchCV(SVC(), tuned_parameters, cv=5, scoring='precision')
# Best found SVC after grid search
# SVC(C=1, break_ties=False, cache_size=200, class_weight=None, coef0=0.0, decision_function_shape='ovr', degree=3, gamma=0.1, kernel='rbf', max_iter=-1, probability=True, random_state=None, shrinking=True, tol=0.001)
model = make_pipeline(StandardScaler(), SVC)
mdl = model.fit(X_train, y_train)
\end{lstlisting}

We have found that the usage of \texttt{StandardScalar} in the model pipeline is causing the model BM5 to be fairer. 
Especially DI of BM5 is 0.14 whereas, the mean of other seven BM models is very high (0.74). \texttt{StandardScalar} transforms the data features independently so that the mean value becomes 0 and the standard deviation becomes 1. Essentially, if a feature has variance in orders of magnitude than another feature, the model might learn from the dominating feature more, which might not be desirable \cite{svc}. In this case, Bank Marketing dataset has 55 features among which 41 has mean close to 0 ([0, 0.35]). However, \textit{age} is the protected attribute having a mean value 0.97 (\textit{older}: 1, \textit{younger}: 0), since the number of older is significantly more than younger. Therefore, \textit{age} is the dominating feature in these BM models. BM5 mitigates that effect by using standard scaling to all features. Therefore, balancing the importance of protected feature with other features can help to reduce bias in the models. 
This example also shows the importance of understanding the underlying properties of protected features and their effectiveness on prediction.\\

\finding{Dropping a feature from the dataset can change the model fairness effectively.}
Both the models AC5 and AC6 are using XGB classifier for prediction but AC6 is fairer than AC5. Among the metrics, in terms of consistency (CNT), AC5 shows bias 3.61 times more than AC6. We have investigated the model construction and found that AC5 and AC6 differ in three constructs: features used in the model, number of trees used in the random forest, and learning rate of the classifier. We have observed that the number of trees and learning rate did not change the bias of the models. In AC5, the model excluded one feature from the training data. 
Bank Marketing dataset contains personal information about individuals and predicts whether the person has an annual income more than 50K dollars or not. In AC5, the model developer dropped one feature that contains number of years of education, since there is other categorical feature which represents education of the person (e.g., bachelors, doctorate, etc.). AC6 is using all the features in the dataset. CNT measures the individual fairness of the models i.e., how two similar individuals (not necessarily from different groups of protected attribute class) are classified to different outcomes. Therefore, dropping the number of years of education is causing the model to classify similar individuals to different outcome, which in turn generating individual unfairness. \\

\begin{figure}[!h]
	\centering
	\includegraphics[width=.83\columnwidth]{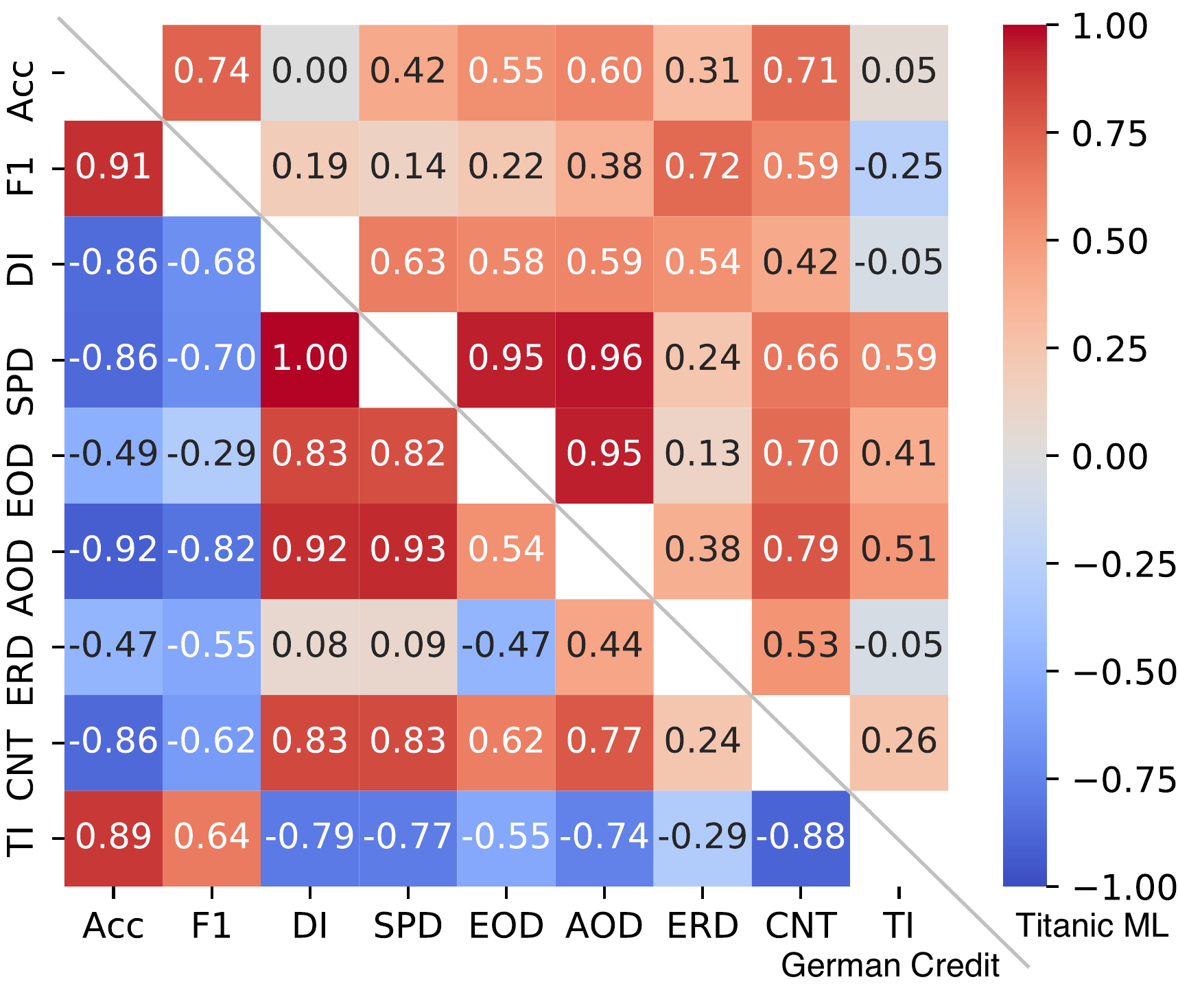}
	\caption{Corelation between the metrics. Bottom diagonal is for German Credit models, top diagonal is for Titanic ML models.}
	\label{corelation}
\end{figure}

\input{main-table}
\finding{Different metrics are needed to understand bias in different models.}
From \fignref{unfairness}, we can see that the models show different patterns of bias in terms of different fairness metrics. For example, compared to any Bank Marketing models, BM5 has disparity impact (DI) less than half but the error rate difference (ERD) more than twice. If the model developer only accounts for DI, then the model would appear fairer than what it actually is. As another example, GC6 is fairer than 90\% of all the models in terms of total bias but if we only consider consistency (CNT), GC6 is fairer than only 50\% of all the models. 
However, previous studies show that achieving fairness with respect to all the metrics is difficult and for some pair of metrics, mathematically impossible \cite{kleinberg2016inherent, chouldechova2017fair, berk2018fairness}. Also, the definition of fairness can vary depending on the application context and the stakeholders. Therefore, it is important to report on comprehensive set of fairness measures and evaluate the trade-off between the metrics to build fairer. 
We have plotted the correlation between different metrics from two datasets in \fignref{corelation}.
A few metric pairs have a similar correlation in both the datasets such as (SPD, EOD), (SPD, AOD). This is understandable from the definitions of these metrics because they are calculated using same or correlated group conditioned rates (true-positives and false-positives). Although there are many metric pairs which are positively or negatively correlated, there is no pattern in correlation values between the two datasets. For instance, CNT and TI are highly negatively correlated in German Credit models but positively correlated in Titanic ML models. Therefore, we need a comprehensive set of metrics to evaluate fairness. \\

\begin{figure}[!h]
	\centering
	\includegraphics[width=\columnwidth]{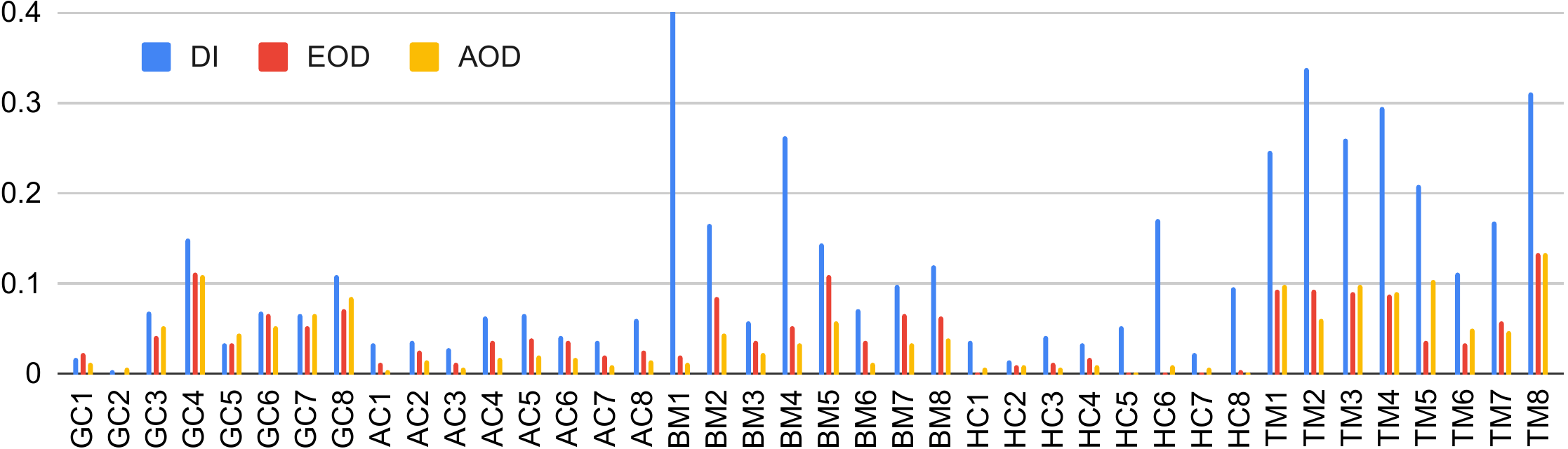}
	\caption{Standard deviation of the metrics: DI, EOD and AOD over multiple experiments. Other metrics have very low standard deviation.}
	\label{std}
\end{figure}

\finding{Except DI, EOD, and AOD, all the fairness measures remain consistent over multiple training and prediction.}
To measure the stability of the fairness and performance metrics, we have computed the standard deviation of each metric over 10 runs similar to \cite{friedler2019comparative}. In each run, the dataset is shuffled before the train-test split, and model is trained on a new randomized training set. We have seen that the models are stable for the performance metrics and most of the fairness metrics. In particular, the average of the standard deviations of accuracy, F1 score, DI, SPD, EOD, AOD, ERD, CNT and TI over all the models are 0.01, 0.01, 0.12, 0.03, 0.04, 0.04, 0.03, 0.01, 0.01, respectively. Except for DI, EOD and AOD, the average standard deviation is very low (less than 0.03). For these three metrics, we have plotted the standard deviations in \fignref{std}. We can see that the trend of standard deviations is similar to the models of a specific dataset. In our benchmark, the largest dataset is Home Credit, which has the lowest standard deviation and the smallest dataset is Titanic ML, which has the most. Since in larger dataset, even after shuffling the training data remains more consistent, the deviation is less. On the other hand, the Titanic ML dataset is the smallest in size, having 891 data instances. The class distribution of data instances do not remain consistent when a random training set is chosen. Therefore, while dealing with smaller datasets, it is important to choose a training set that represents the original data and evaluate fairness multiple times. \\

DI has more standard deviation than other metrics. DI is computed using the ratio of two probabilities, $P_u/P_p$, where $P_u$ is the probability of unprivileged group getting favorable label, and $P_p$ is the probability of privileged group getting favorable label. Even the probability difference is very low, the value of DI can be very high. Therefore, DI fluctuates more frequently than other metrics.

\begin{figure}[!h]
	\centering
	\includegraphics[width=\columnwidth]{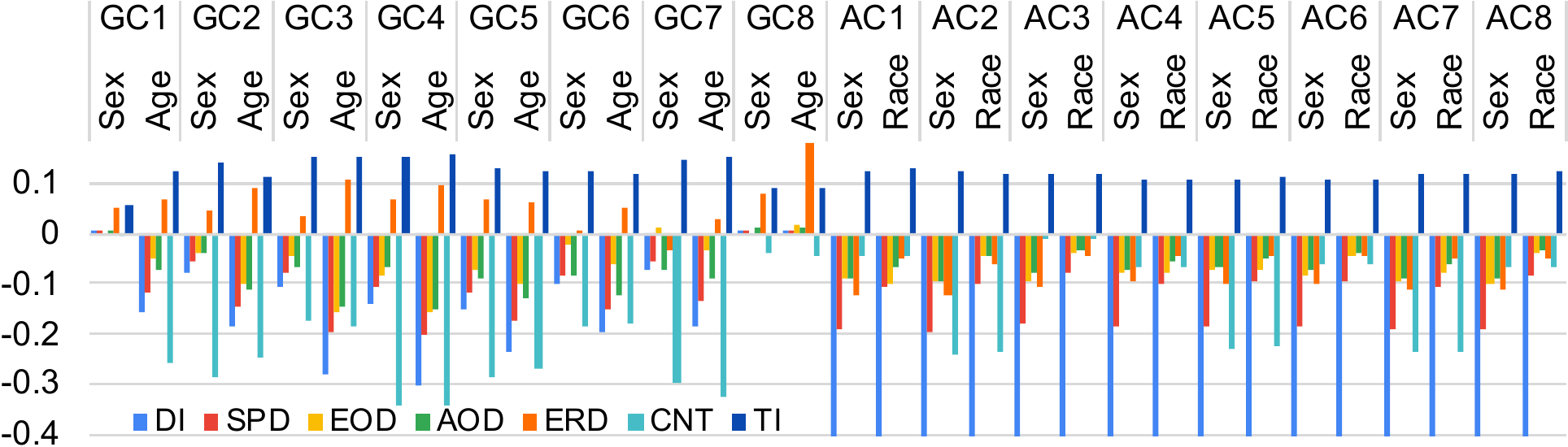}
	\caption{Fairness of ML models with respect to different protected attributes}
	\label{multi-pro}
\end{figure}

\finding{A fair model with respect to one protected attribute is not necessarily fair with respect to another protected attribute.}
To understand the behavior of the same models on different protected attributes, we have analyzed the fairness of German Credit and Adult Census models on two protected attributes. In \fignref{multi-pro}, we have plotted the fairness measures of German Credit models on \textit{sex} and \textit{age} and Adult Census models on \textit{sex} and \textit{race}. We have found that the models can show different fairness when different protected attribute is considered. The total bias exhibited by German Credit dataset are: for \textit{sex} attribute 4.82 and for \textit{age} attribute 7.72. For Adult Census, the total bias are: for \textit{sex} attribute 15.15 and for \textit{race} attribute 8.56. However, most of the models exhibit similar trend of difference in the fairness when considering two different attributes.

GC1 and GC6 show cumulative bias 0.12 and 0.60 when \textit{sex} is considered. Surprisingly, GC1 and GC6 shows cumulative bias 0.85 and 0.88 when \textit{age} is considered. GC1 is much fairer model than GC6 in the first case but in the second case, the fairness is almost similar. We have discussed the behavior of these two models in Finding 1 and explained how GC1 is fairer when \textit{sex} is the protected attribute. However, the fair prediction does not persist for the \textit{age} because there is no imbalance in German Credit with respect to \textit{age} groups. Therefore, GC1 and GC6 show similar fairness when \textit{age} is considered. 

%% file: main-table.tex
\begin{table*}[!h]
	\vspace{1.3mm}
	\footnotesize
	\setlength\tabcolsep{4.3pt} 
	\centering
	\caption{Unfairness measures of the models before and after the mitigations}
	\label{tab:main}%
	\begin{tabular}{|r|l|r|r|r|r|r|r|r|r|r|r|r|r|r|r|r|r|r|r|
			>{\columncolor[HTML]{EFEFEF}}l |}
		\hline
		& \multicolumn{1}{c|}{}                                 & \multicolumn{9}{c|}{\cellcolor[HTML]{EFEFEF}\textbf{Before mitigation}}                                                                                                                                                                                                                                                                                   & \multicolumn{10}{c|}{\cellcolor[HTML]{EFEFEF}\textbf{After mitigation}}                                                                                                                                                                                                                                                                                                                                                \\ \hhline{|~|~|*{19}{-}}
		\multirow{-2}{*}{} & \multicolumn{1}{c|}{\multirow{-2}{*}{\textbf{Model}}} & \cellcolor[HTML]{EFEFEF}\textbf{Acc} & \cellcolor[HTML]{EFEFEF}\textbf{F1} & \cellcolor[HTML]{EFEFEF}\textbf{DI} & \cellcolor[HTML]{EFEFEF}\textbf{SPD} & \cellcolor[HTML]{EFEFEF}\textbf{EOD} & \cellcolor[HTML]{EFEFEF}\textbf{AOD} & \cellcolor[HTML]{EFEFEF}\textbf{ERD} & \cellcolor[HTML]{EFEFEF}\textbf{CNT} & \cellcolor[HTML]{EFEFEF}\textbf{TI} & \cellcolor[HTML]{EFEFEF}\textbf{Acc} & \cellcolor[HTML]{EFEFEF}\textbf{F1} & \cellcolor[HTML]{EFEFEF}\textbf{DI} & \cellcolor[HTML]{EFEFEF}\textbf{SPD} & \cellcolor[HTML]{EFEFEF}\textbf{EOD} & \cellcolor[HTML]{EFEFEF}\textbf{AOD} & \cellcolor[HTML]{EFEFEF}\textbf{ERD} & \cellcolor[HTML]{EFEFEF}\textbf{CNT} & \cellcolor[HTML]{EFEFEF}\textbf{TI} & \multicolumn{1}{c|}{\cellcolor[HTML]{EFEFEF}\textbf{Rank}} \\ \hline
		& GC1-RFT                                               & .687                                 & .814                                & .002                                & .002                                 & 0                                 & .004                                 & .052                                 & -.002                                & .058                                & \cellcolor{cellblue}.683         & \cellcolor{cellblue}.811        & \cellcolor{cellblue}.002        & \cellcolor{cellblue}.002         & \cellcolor{cellblue}0         & \cellcolor{cellblue}.004         & \cellcolor{cellblue}-.032        & \cellcolor{cellblue}-.002        & \cellcolor{cellblue}.058        & RAOD/PCE                                                   \\ \hhline{|~|*{20}{-}}
		& GC2-XGB                                               & .743                                 & .828                                & -.076                               & -.058                                & -.039                                & -.036                                & .047                                 & -.282                                & .142                                & \cellcolor{cellgreen}.709         & \cellcolor{cellgreen}.829        & \cellcolor{cellgreen}0        & \cellcolor{cellgreen}0         & \cellcolor{cellgreen}0         & \cellcolor{cellgreen}0         & \cellcolor{cellgreen}.067         & \cellcolor{cellgreen}0         & \cellcolor{cellgreen}.057        & AORD/PCE                                                   \\ \hhline{|~|*{20}{-}}
		& GC3-XGB                                               & .742                                 & .827                                & -.105                               & -.079                                & -.043                                & -.065                                & .036                                 & -.173                                & .149                                & \cellcolor{cellgreen}.729         & \cellcolor{cellgreen}.831        & \cellcolor{cellgreen}-.045       & \cellcolor{cellgreen}-.040        & \cellcolor{cellgreen}-.006        & \cellcolor{cellgreen}-.043        & \cellcolor{cellgreen}.037         & \cellcolor{cellgreen}-.095        & \cellcolor{cellgreen}.100        & AR/DPOCE                                                   \\ \hhline{|~|*{20}{-}}
		& GC4-SVC                                               & .753                                 & .832                                & -.138                               & -.104                                & -.081                                & -.068                                & .070                                 & -.338                                & .153                                & \cellcolor{cellgreen}.716         & \cellcolor{cellgreen}.834        & \cellcolor{cellgreen}0        & \cellcolor{cellgreen}0         & \cellcolor{cellgreen}0         & \cellcolor{cellgreen}0         & \cellcolor{cellgreen}.090         & \cellcolor{cellgreen}0         & \cellcolor{cellgreen}.057        & AORD/PEC                                                   \\ \hhline{|~|*{20}{-}}
		& GC5-EVC                                               & .743                                 & .826                                & -.148                               & -.116                                & -.075                                & -.089                                & .067                                 & -.286                                & .127                                & \cellcolor{cellgreen}.687         & \cellcolor{cellgreen}.814        & \cellcolor{cellgreen}0        & \cellcolor{cellgreen}0         & \cellcolor{cellgreen}0         & \cellcolor{cellgreen}0         & \cellcolor{cellgreen}.112         & \cellcolor{cellgreen}0         & \cellcolor{cellgreen}.058        & AORD/PEC                                                   \\ \hhline{|~|*{20}{-}}
		& GC6-RFT                                               & .761                                 & .845                                & -.103                               & -.083                                & -.023                                & -.085                                & .005                                 & -.183                                & .121                                & \cellcolor{cellblue}.759         & \cellcolor{cellblue}.844        & \cellcolor{cellblue}-.071       & \cellcolor{cellblue}-.058        & \cellcolor{cellblue}-.023        & \cellcolor{cellblue}-.085        & \cellcolor{cellblue}-.027        & \cellcolor{cellblue}-.183        & \cellcolor{cellblue}.121        & RD/APCEO                                                   \\ \hhline{|~|*{20}{-}}
		& GC7-XGB                                               & .751                                 & .831                                & -.073                               & -.056                                & .009                                 & -.072                                & -.033                                & -.293                                & .144                                & \cellcolor{cellgreen}.709         & \cellcolor{cellgreen}.829        & \cellcolor{cellgreen}0        & \cellcolor{cellgreen}0         & \cellcolor{cellgreen}0         & \cellcolor{cellgreen}0         & \cellcolor{cellgreen}.047         & \cellcolor{cellgreen}0         & \cellcolor{cellgreen}.057        & ADR/POCE                                                   \\ \hhline{|~|*{20}{-}}
		\multirow{-8}{*}{\rotatebox{90}{\shortstack{German Credit (Sex)*}}}            & GC8-KNN                                               & .698                                 & .815                                & .003                                & .002                                 & 0                                 & .011                                 & .081                                 & -.041                                & .090                                & \cellcolor{cellgreen}.702         & \cellcolor{cellgreen}.825        & \cellcolor{cellgreen}0        & \cellcolor{cellgreen}0         & \cellcolor{cellgreen}0         & \cellcolor{cellgreen}0         & \cellcolor{cellgreen}.086         & \cellcolor{cellgreen}0         & \cellcolor{cellgreen}.057        & AR/DPCOE                                                   \\ \hline
		& AC1-LRG                                               & .845                                 & .657                                & -.654                               & -.104                                & -.100                                & -.069                                & -.050                                & -.045                                & .127                                & \cellcolor{cellorange}.261         & \cellcolor{cellorange}.399        & \cellcolor{cellorange}.023        & \cellcolor{cellorange}.023         & \cellcolor{cellorange}.017         & \cellcolor{cellorange}.021         & \cellcolor{cellorange}.120         & \cellcolor{cellorange}-.019        & \cellcolor{cellorange}.040        & ORCDAP/E                                                   \\ \hhline{|~|*{20}{-}}
		& AC2-RFT                                               & .846                                 & .657                                & -.582                               & -.098                                & -.047                                & -.046                                & -.060                                & -.236                                & .119                                & \cellcolor{cellgreen}.787         & \cellcolor{cellgreen}.249        & \cellcolor{cellgreen}-.354       & \cellcolor{cellgreen}-.014        & \cellcolor{cellgreen}.007         & \cellcolor{cellgreen}.003         & \cellcolor{cellgreen}-.086        & \cellcolor{cellgreen}-.005        & \cellcolor{cellgreen}.232        & AROC/DPE                                                   \\ \hhline{|~|*{20}{-}}
		& AC3-GBC                                               & .858                                 & .677                                & -.496                               & -.079                                & -.041                                & -.031                                & -.045                                & -.010                                & .120                                & \cellcolor{cellblue}.858         & \cellcolor{cellblue}.675        & \cellcolor{cellblue}-.131       & \cellcolor{cellblue}-.024        & \cellcolor{cellblue}-.041        & \cellcolor{cellblue}-.031        & \cellcolor{cellblue}-.004        & \cellcolor{cellblue}-.010        & \cellcolor{cellblue}.120        & ROAC/DPE                                                   \\ \hhline{|~|*{20}{-}}
		& AC4-CBC                                               & .869                                 & .712                                & -.616                               & -.102                                & -.077                                & -.056                                & -.044                                & -.069                                & .107                                & \cellcolor{cellorange}.805         & \cellcolor{cellorange}.683        & \cellcolor{cellorange}-.127       & \cellcolor{cellorange}-.044        & \cellcolor{cellorange}.080         & \cellcolor{cellorange}.044         & \cellcolor{cellorange}-.001        & \cellcolor{cellorange}-.102        & \cellcolor{cellorange}.082        & ORAC/PDE                                                   \\ \hhline{|~|*{20}{-}}
		& AC5-XGB                                               & .867                                 & .708                                & -.588                               & -.097                                & -.073                                & -.051                                & -.043                                & -.224                                & .111                                & \cellcolor{cellblue}.865         & \cellcolor{cellblue}.705        & \cellcolor{cellblue}-.203       & \cellcolor{cellblue}-.039        & \cellcolor{cellblue}-.073        & \cellcolor{cellblue}-.051        & \cellcolor{cellblue}-.002        & \cellcolor{cellblue}-.224        & \cellcolor{cellblue}.111        & ROAC/PDE                                                   \\ \hhline{|~|*{20}{-}}
		& AC6-XGB                                               & .871                                 & .717                                & -.570                               & -.096                                & -.044                                & -.036                                & -.047                                & -.062                                & .106                                & \cellcolor{cellorange}.808         & \cellcolor{cellorange}.691        & \cellcolor{cellorange}-.132       & \cellcolor{cellorange}-.046        & \cellcolor{cellorange}.072         & \cellcolor{cellorange}.044         & \cellcolor{cellorange}.009         & \cellcolor{cellorange}-.094        & \cellcolor{cellorange}.078        & ORAC/PDE                                                   \\ \hhline{|~|*{20}{-}}
		& AC7-RFT                                               & .852                                 & .678                                & -.615                               & -.104                                & -.078                                & -.059                                & -.051                                & -.235                                & .117                                & \cellcolor{cellgreen}.638         & \cellcolor{cellgreen}.329        & \cellcolor{cellgreen}-.289       & \cellcolor{cellgreen}-.024        & \cellcolor{cellgreen}-.005        & \cellcolor{cellgreen}-.009        & \cellcolor{cellgreen}-.039        & \cellcolor{cellgreen}-.009        & \cellcolor{cellgreen}.187        & AORCD/PE                                                   \\ \hhline{|~|*{20}{-}}
		\multirow{-8}{*}{\rotatebox{90}{\shortstack{Adult Census (Race)*}}}            & AC8-DCT                                               & .853                                 & .675                                & -.519                               & -.086                                & -.040                                & -.035                                & -.050                                & -.068                                & .121                                & \cellcolor{cellblue}.852         & \cellcolor{cellblue}.673        & \cellcolor{cellblue}-.153       & \cellcolor{cellblue}-.029        & \cellcolor{cellblue}-.040        & \cellcolor{cellblue}-.035        & \cellcolor{cellblue}-.010        & \cellcolor{cellblue}-.068        & \cellcolor{cellblue}.121        & ROAC/DPE                                                   \\ \hline
		& BM1-XGB                                               & .906                                 & .582                                & .627                                & .087                                 & .074                                 & .053                                 & .051                                 & -.078                                & .074                                & \cellcolor{cellblue}.905         & \cellcolor{cellblue}.581        & \cellcolor{cellblue}.274        & \cellcolor{cellblue}.032         & \cellcolor{cellblue}.074         & \cellcolor{cellblue}.053         & \cellcolor{cellblue}.017         & \cellcolor{cellblue}-.078        & \cellcolor{cellblue}.074        & ROCPD/EA                                                   \\ \hhline{|~|*{20}{-}}
		& BM2-LGB                                               & .908                                 & .606                                & .593                                & .083                                 & .004                                 & .022                                 & .069                                 & -.034                                & .072                                & \cellcolor{cellorange}.772         & \cellcolor{cellorange}.498        & \cellcolor{cellorange}.076        & \cellcolor{cellorange}.026         & \cellcolor{cellorange}-.037        & \cellcolor{cellorange}-.037        & \cellcolor{cellorange}-.031        & \cellcolor{cellorange}-.040        & \cellcolor{cellorange}.066        & ORDC/PAE                                                   \\ \hhline{|~|*{20}{-}}
		& BM3-GBC                                               & .908                                 & .604                                & .688                                & .100                                 & .083                                 & .056                                 & .051                                 & -.032                                & .072                                & \cellcolor{cellred}.852         & \cellcolor{cellred}.529        & \cellcolor{cellred}.066        & \cellcolor{cellred}.013         & \cellcolor[HTML]{FFCCCC}-.059        & \cellcolor[HTML]{FFCCCC}-.052        & \cellcolor{cellred}006         & \cellcolor[HTML]{FFCCCC}-.089        & \cellcolor{cellred}078        & CODR/APE                                                   \\ \hhline{|~|*{20}{-}}
		& BM4-XGB                                               & .887                                 & .330                                & .810                                & .048                                 & .067                                 & .042                                 & .074                                 & -.010                                & .111                                & \cellcolor{cellblue}.887         & \cellcolor{cellblue}.328        & \cellcolor{cellblue}.442        & \cellcolor{cellblue}.022         & \cellcolor{cellblue}.067         & \cellcolor{cellblue}.042         & \cellcolor{cellblue}.001         & \cellcolor{cellblue}-.010        & \cellcolor{cellblue}.111        & RCA/OPDE                                                   \\ \hhline{|~|*{20}{-}}
		& BM5-SVC                                               & .875                                 & .175                                & .139                                & .003                                 & -.077                                & -.031                                & .126                                 & -.032                                & .126                                & \cellcolor{cellpurple}.873         & \cellcolor{cellpurple}.002        & \cellcolor{cellpurple}.139        & \cellcolor{cellpurple}0         & \cellcolor{cellpurple}-.001        & \cellcolor{cellpurple}0         & \cellcolor{cellpurple}.110         & \cellcolor{cellpurple}0         & \cellcolor{cellpurple}.136        & ERCDO/AP                                                   \\ \hhline{|~|*{20}{-}}
		& BM6-GBC                                               & .908                                 & .612                                & .698                                & .105                                 & .030                                 & .038                                 & .076                                 & -.033                                & .071                                & \cellcolor{cellorange}.795         & \cellcolor{cellorange}.521        & \cellcolor{cellorange}.110        & \cellcolor{cellorange}.034         & \cellcolor{cellorange}-.072        & \cellcolor{cellorange}-.053        & \cellcolor{cellorange}-.019        & \cellcolor{cellorange}-.039        & \cellcolor{cellorange}.065        & OCRD/PAE                                                   \\ \hhline{|~|*{20}{-}}
		& BM7-XGB                                               & .910                                 & .611                                & .713                                & .107                                 & .051                                 & .052                                 & .072                                 & -.047                                & .070                                & \cellcolor{cellred}.829         & \cellcolor{cellred}.485        & \cellcolor{cellred}.022        & \cellcolor{cellred}.004         & \cellcolor[HTML]{FFCCCC}-.037        & \cellcolor[HTML]{FFCCCC}-.044        & \cellcolor[HTML]{FFCCCC}-.007        & \cellcolor[HTML]{FFCCCC}-.122        & \cellcolor{cellred}.085        & CODRA/PE                                                   \\ \hhline{|~|*{20}{-}}
		\multirow{-8}{*}{\rotatebox{90}{\shortstack{Bank Marketing (Age)}}}            & BM8-RFT                                               & .899                                 & .435                                & .834                                & .066                                 & .091                                 & .058                                 & .064                                 & -.023                                & .097                                & \cellcolor{cellorange}.795         & \cellcolor{cellorange}.462        & \cellcolor{cellorange}.289        & \cellcolor{cellorange}.042         & \cellcolor{cellorange}-.048        & \cellcolor{cellorange}-.027        & \cellcolor{cellorange}.005         & \cellcolor{cellorange}-.052        & \cellcolor{cellorange}.073        & ORACDP/E                                                   \\ \hline
		& HC1-LGB                                               & .883                                 & .249                                & .574                                & .046                                 & .065                                 & .052                                 & .051                                 & -.110                                & .083                                & \cellcolor{cellgreen}.238         & \cellcolor{cellgreen}.132        & \cellcolor{cellgreen}-.025       & \cellcolor{cellgreen}-.002        & \cellcolor{cellgreen}-.003        & \cellcolor{cellgreen}-.002        & \cellcolor{cellgreen}-.020        & \cellcolor{cellgreen}-.006        & \cellcolor{cellgreen}.030        & APECR/OD                                                   \\ \hhline{|~|*{20}{-}}
		& HC2-LGB                                               & .920                                 & .094                                & -.698                               & -.006                                & -.016                                & -.010                                & -.032                                & -.012                                & .081                                & \cellcolor{cellgold}.919         & \cellcolor{cellgold}.002        & \cellcolor{cellgold}.076        & \cellcolor{cellgold}0         & \cellcolor{cellgold}0         & \cellcolor{cellgold}0         & \cellcolor{cellgold}-.033        & \cellcolor{cellgold}0         & \cellcolor{cellgold}.084        & PECROA/D                                                   \\ \hhline{|~|*{20}{-}}
		& HC3-GNB                                               & .913                                 & .010                                & .974                                & .999                                 & .007                                 & .005                                 & .006                                 & -2.449                               & 0                                & \cellcolor{cellorange}.732         & \cellcolor{cellorange}.194        & \cellcolor{cellorange}.181        & \cellcolor{cellorange}.857         & \cellcolor{cellorange}.047         & \cellcolor{cellorange}.019         & \cellcolor{cellorange}.031         & \cellcolor{cellorange}-2.285       & \cellcolor{cellorange}0        & OA/DECPR                                                   \\ \hhline{|~|*{20}{-}}
		& HC4-XGB                                               & .919                                 & .046                                & .868                                & .994                                 & .003                                 & .013                                 & .007                                 & -2.482                               & 0                                & \cellcolor{cellred}.918         & \cellcolor{cellred}.012        & \cellcolor[HTML]{FFCCCC}-.103       & \cellcolor{cellred}.998         & \cellcolor{cellred}0         & \cellcolor[HTML]{FFCCCC}-.003        & \cellcolor[HTML]{FFCCCC}-.002        & \cellcolor[HTML]{FFCCCC}-2.468       & \cellcolor{cellred}0        & CEDRP/OA                                                   \\ \hhline{|~|*{20}{-}}
		& HC5-CBC                                               & .870                                 & .302                                & .744                                & .865                                 & .085                                 & .140                                 & .106                                 & -2.524                               & 0                                & \cellcolor{cellgreen}.552         & \cellcolor{cellgreen}.075        & \cellcolor{cellgreen}-.134       & \cellcolor{cellgreen}.999         & \cellcolor{cellgreen}-.025        & \cellcolor{cellgreen}-.017        & \cellcolor{cellgreen}-.021        & \cellcolor{cellgreen}-2.772       & \cellcolor{cellgreen}.001        & ACEPR/DO                                                   \\ \hhline{|~|*{20}{-}}
		& HC6-CBC                                               & .869                                 & .305                                & .735                                & .085                                 & .144                                 & .107                                 & .068                                 & -.147                                & .080                                & \cellcolor{cellgreen}.583         & \cellcolor{cellgreen}.074        & \cellcolor{cellgreen}.021        & \cellcolor{cellgreen}0         & \cellcolor{cellgreen}0         & \cellcolor{cellgreen}0         & \cellcolor{cellgreen}.007         & \cellcolor{cellgreen}0         & \cellcolor{cellgreen}.056        & ACPER/DO                                                   \\ \hhline{|~|*{20}{-}}
		& HC7-XGB                                               & .911                                 & .211                                & .953                                & .953                                 & .033                                 & .084                                 & .054                                 & -2.533                               & 0                                & \cellcolor{cellpurple}.907         & \cellcolor{cellpurple}.090        & \cellcolor{cellpurple}.408        & \cellcolor{cellpurple}.966         & \cellcolor{cellpurple}.009         & \cellcolor{cellpurple}-.052        & \cellcolor{cellpurple}-.019        & \cellcolor{cellpurple}-2.453       & \cellcolor{cellpurple}0        & ECPR/DOA                                                   \\ \hhline{|~|*{20}{-}}
		\multirow{-8}{*}{\rotatebox{90}{\shortstack{Home Credit (Sex)}}}            & HC8-RFT                                               & .661                                 & .239                                & .383                                & .719                                 & .147                                 & .129                                 & .133                                 & -2.449                               & .001                                & \cellcolor{cellred}.645         & \cellcolor{cellred}.226        & \cellcolor{cellred}.337        & \cellcolor{cellred}.681         & \cellcolor{cellred}.133         & \cellcolor{cellred}.098         & \cellcolor{cellred}.112         & \cellcolor[HTML]{FFCCCC}-2.426       & \cellcolor{cellred}.001        & CPRD/AEO                                                   \\ \hline
		& TM1-XGB                                               & .807                                 & .720                                & -2.247                              & -.705                                & -.631                                & -.559                                & -.056                                & -.341                                & .153                                & \cellcolor{cellorange}.649         & \cellcolor{cellorange}.580        & \cellcolor{cellorange}-.082       & \cellcolor{cellorange}-.039        & \cellcolor{cellorange}.027         & \cellcolor{cellorange}.177         & \cellcolor{cellorange}.115         & \cellcolor{cellorange}-.272        & \cellcolor{cellorange}.189        & OAERDP/C                                                   \\ \hhline{|~|*{20}{-}}
		& TM2-RFT                                               & .816                                 & .753                                & -2.013                              & -.709                                & -.635                                & -.515                                & .022                                 & -.293                                & .142                                & \cellcolor{cellorange}.644         & \cellcolor{cellorange}.566        & \cellcolor{cellorange}-.106       & \cellcolor{cellorange}-.045        & \cellcolor{cellorange}.059         & \cellcolor{cellorange}.166         & \cellcolor{cellorange}.023         & \cellcolor{cellorange}-.269        & \cellcolor{cellorange}.223        & OAERDP/C                                                   \\ \hhline{|~|*{20}{-}}
		& TM3-EBG                                               & .799                                 & .725                                & -2.125                              & -.674                                & -.637                                & -.514                                & -.017                                & -.333                                & .165                                & \cellcolor{cellorange}.647         & \cellcolor{cellorange}.572        & \cellcolor{cellorange}-.108       & \cellcolor{cellorange}-.045        & \cellcolor{cellorange}.031         & \cellcolor{cellorange}.148         & \cellcolor{cellorange}.050         & \cellcolor{cellorange}-.317        & \cellcolor{cellorange}.223        & OAERD/PC                                                   \\ \hhline{|~|*{20}{-}}
		& TM4-LRG                                               & .800                                 & .732                                & -2.439                              & -.808                                & -.729                                & -.694                                & -.051                                & -.381                                & .144                                & \cellcolor{cellorange}.658         & \cellcolor{cellorange}.577        & \cellcolor{cellorange}-.075       & \cellcolor{cellorange}-.034        & \cellcolor{cellorange}.072         & \cellcolor{cellorange}.160         & \cellcolor{cellorange}.038         & \cellcolor{cellorange}-.327        & \cellcolor{cellorange}.207        & OAEPRD/C                                                   \\ \hhline{|~|*{20}{-}}
		& TM5-GBC                                               & .816                                 & .740                                & -2.268                              & -.708                                & -.647                                & -.542                                & -.022                                & -.357                                & .151                                & \cellcolor{cellorange}.651         & \cellcolor{cellorange}.572        & \cellcolor{cellorange}-.087       & \cellcolor{cellorange}-.033        & \cellcolor{cellorange}.097         & \cellcolor{cellorange}.174         & \cellcolor{cellorange}.029         & \cellcolor{cellorange}-.332        & \cellcolor{cellorange}.205        & OAERD/CP                                                   \\ \hhline{|~|*{20}{-}}
		& TM6-XGB                                               & .804                                 & .730                                & -1.948                              & -.665                                & -.583                                & -.499                                & -.042                                & -.345                                & .146                                & \cellcolor{cellorange}.625         & \cellcolor{cellorange}.568        & \cellcolor{cellorange}-.079       & \cellcolor{cellorange}-.038        & \cellcolor{cellorange}.075         & \cellcolor{cellorange}.157         & \cellcolor{cellorange}.092         & \cellcolor{cellorange}-.367        & \cellcolor{cellorange}.190        & OAERD/CP                                                   \\ \hhline{|~|*{20}{-}}
		& TM7-RFT                                               & .825                                 & .747                                & -2.232                              & -.639                                & -.555                                & -.411                                & -.029                                & -.285                                & .161                                & \cellcolor{cellorange}.653         & \cellcolor{cellorange}.577        & \cellcolor{cellorange}-.099       & \cellcolor{cellorange}-.043        & \cellcolor{cellorange}.100         & \cellcolor{cellorange}.188         & \cellcolor{cellorange}.003         & \cellcolor{cellorange}-.261        & \cellcolor{cellorange}.219        & OAERDP/C                                                   \\ \hhline{|~|*{20}{-}}
		\multirow{-8}{*}{\rotatebox{90}{\shortstack{Titanic ML (Sex)}}}            & TM8-RFT                                               & .814                                 & .732                                & -2.306                              & -.716                                & -.633                                & -.563                                & -.051                                & -.321                                & .149                                & \cellcolor{cellorange}.649         & \cellcolor{cellorange}.596        & \cellcolor{cellorange}-.082       & \cellcolor{cellorange}-.042        & \cellcolor{cellorange}.011         & \cellcolor{cellorange}.166         & \cellcolor{cellorange}.157         & \cellcolor{cellorange}-.327        & \cellcolor{cellorange}.172        & OAERD/PC                                                   \\ \hline
	\end{tabular}
	\\
*Experiment has been conducted for multiple protected attributes. RFT: Random Forest, XGB: XGBoost, SVC: Support Vector Classifier, EVC: Ensemble Voting Classifier, KNN: K-Nearest Neighbors, LRG: Logistic Regression, GBC: Gradient Boosting Classifier, CBC: Cat Boost Classifier, DCT: Decision Tree, LGB: Light Gradient Boost, GNB: Gaussian Naive Bayes, EBG: Ensemble Bagging. Mitigation techniques applied to the models are as follows. Result is shown for the best mitigation. Rank of mitigation uses acronym below (mitigations before `/' have been able to mitigate bias, rest have not.) \\
\colorbox{cellblue}{Reweighing (R)} \colorbox{cellsteel}{DI Remover (D)} \colorbox{cellgreen}{Adversarial Debiasing (A)} \colorbox{cellgold}{Prejudice Remover (P)} \colorbox{cellpurple}{Equalized Odds(E)} \colorbox{cellred}{Calibrated Equalized Odds (C)} \colorbox{cellorange}{Reject Option Classification (O)}
\end{table*}%

%% file: mitigation.tex
\section{Mitigation}
\label{sec:mitigation}
\begin{figure*}[!t]
	\centering
	\includegraphics[width=\linewidth]{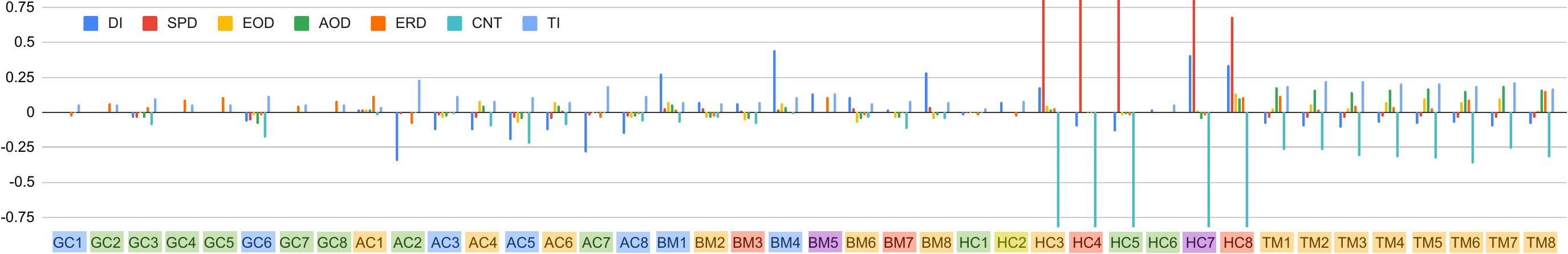}
	\caption{The fairness exhibited by the models after applying the bias mitigation techniques. The color coding in \tabref{tab:main} is used to denote the most successful mitigation algorithm for each model.}
	\label{fairness}
\end{figure*}

In this section, we have investigated the fairness results of the models after applying bias mitigation techniques. We have employed 7 different bias mitigation algorithms separately on 40 models and compared the fairness results with the original fairness exhibited by the models. For each model, we have selected the most successful mitigation algorithm and plotted the fairness values after mitigation in \fignref{fairness}. We have observed that similar to \fignref{unfairness}, the fairness patterns are similar for the models in a dataset. DI, SPD, and CNT are the most difficult metrics to mitigate.

To understand the root causes of unfairness, we have focused on the models which exhibit more or less bias and then investigated the effects of different mitigation algorithms.
Here, among the mitigation algorithms, the preprocessing techniques operate on the training data and retrain the original model to remove bias. On the other hand, post-processing techniques do not change the training data or original model but change the prediction made by the model. The in-processing techniques do not alter the dataset or prediction result but employ completely new modeling technique. \\

\finding{Models with effective preprocessing mitigation technique is preferable than others.}
We have found that Reweighing algorithm has effectively debiased many models: GC1, GC6, AC3, AC5, AC8, BM1 and BM4. These models produce fairer results when the dataset is pre-processed using Reweighing. In other words, these models do not propagate bias themselves. In other cases where pre-processing techniques are not effective, we had to change the model or alter the prediction, which implies that bias is induced or propagated by the models. Another advantage is that in these models, after mitigations the models have retained the accuracy and F1 score. Other mitigation techniques often hampered the performance of the model. 
For a few other models (GC3, GC8, AC1, AC2, AC4, AC6, BM2, BM5, BM8), Reweighing has been the second most successful mitigation algorithm. Among these models, in AC1, AC2, BM2, and BM5, the most successful algorithm to mitigate bias loss accuracy or F1 score at least 22\%. In all of these cases, Reweighing has retained both accuracy and F1 score. \\

\finding{Models with more bias are debiased effectively by post-processing techniques, whereas originally fairer models are debiased effectively by preprocessing or in-processing techniques.}
From \tabref{tab:main}, we can see that 21 out of 40 models are debiased by one of the three post-processing algorithms i.e., Equalized odds (EO), Calibrated equalized odds (CEO), and Reject option classifier (ROC). These algorithms have been able to mitigate bias (not necessarily the most successful) in 90\% of the models. Especially, ROC and CEO are the dominant post-processing techniques. ROC takes the model prediction, and gives the favorable outcome to the unprivileged group and unfavorable outcome to privileged group with a certain
confidence around the decision boundary \cite{kamiran2012decision}. CEO takes the probability distribution score generated by the classifier and find the probability of changing outcome label and maximize
equalized odds \cite{pleiss2017fairness}. EO also changes the outcome label with certain probability obtained by solving a linear program \cite{hardt2016equality}. We have found that these post-processing methods have been able to mitigate bias more effectively when the original model produces more biased results. 
From \fignref{cumulative}, we can see that the most biased 5 models are TM4, TM8, TM5, TM1, HC7, where the post-processing has been the most successful algorithms.
On the contrary, in case of the 5 least biased model (GC1, GC8, BM5, GC6, GC3), rather than mitigating, all three post-processing techniques increased bias. 

In \tabref{tab:main}, we have shown the rank of mitigation algorithms to debias each model. 
In \tabref{tab:rank}, we have shown the mean of the ranks of each mitigation algorithms, where rank of most successful algorithm is 1 and least is 7. We can see that for most biased models, Reject option classification and Equalized odds have been more successful than all others. For the least biased models, both preprocessing algorithms and Adversarial Debiasing have been effective, and the post-processing algorithms have been ineffective. \\

\begin{table}[!ht]
	\footnotesize
	\caption{Mean rank of each bias mitigation algorithm for 10 least biased models (LBM), 10 most biased models (MBM), and overall.}
	\label{tab:rank}
	\begin{tabular}{|p{1.5cm}|l|l|l|l|}
		\hline
		\rowcolor[HTML]{C0C0C0} 
		\multicolumn{1}{|l|}{\cellcolor[HTML]{C0C0C0}\textbf{Stage}} & \textbf{Algorithms}                                       & \textbf{LBM}                & \textbf{MBM}                & \textbf{All}              \\ \hline
		& Reweighing (R)                                            & 2.1                           & 4.5                           & 3.03                         \\
		\multirow{-2}{*}{Preprocessing}                              & \cellcolor[HTML]{EFEFEF}Disparate Impact Remover (D)      & \cellcolor[HTML]{EFEFEF}3.7 & \cellcolor[HTML]{EFEFEF}4.8   & \cellcolor[HTML]{EFEFEF}4.58 \\ \hline
		& Adversarial Debiasing (A)                                 & 3                         & 2.9                           & 3                         \\
		\multirow{-2}{*}{In-processing}                              & \cellcolor[HTML]{EFEFEF}Prejudice Remover Regularizer (P) & \cellcolor[HTML]{EFEFEF}4.5 & \cellcolor[HTML]{EFEFEF}5.3   & \cellcolor[HTML]{EFEFEF}4.98 \\ \hline
		& Equalized Odds (E)                                        & 5.8                           & 2.8                           & 5.18                         \\
		& \cellcolor[HTML]{EFEFEF}Calibrated Equalized Odds (C)     & \cellcolor[HTML]{EFEFEF}4.8   & \cellcolor[HTML]{EFEFEF}5.1 & \cellcolor[HTML]{EFEFEF}4.33 \\
		\multirow{-3}{*}{Post-processing}                            & Reject Option Classification (O)                          & 4.1                           & 2.6                           & 2.93 \\ \hline
	\end{tabular}
\end{table}

%% file: impact.tex
\section{Impact}
\label{sec:impact}

\begin{figure}[t]
	\centering
	\includegraphics[width=\columnwidth]{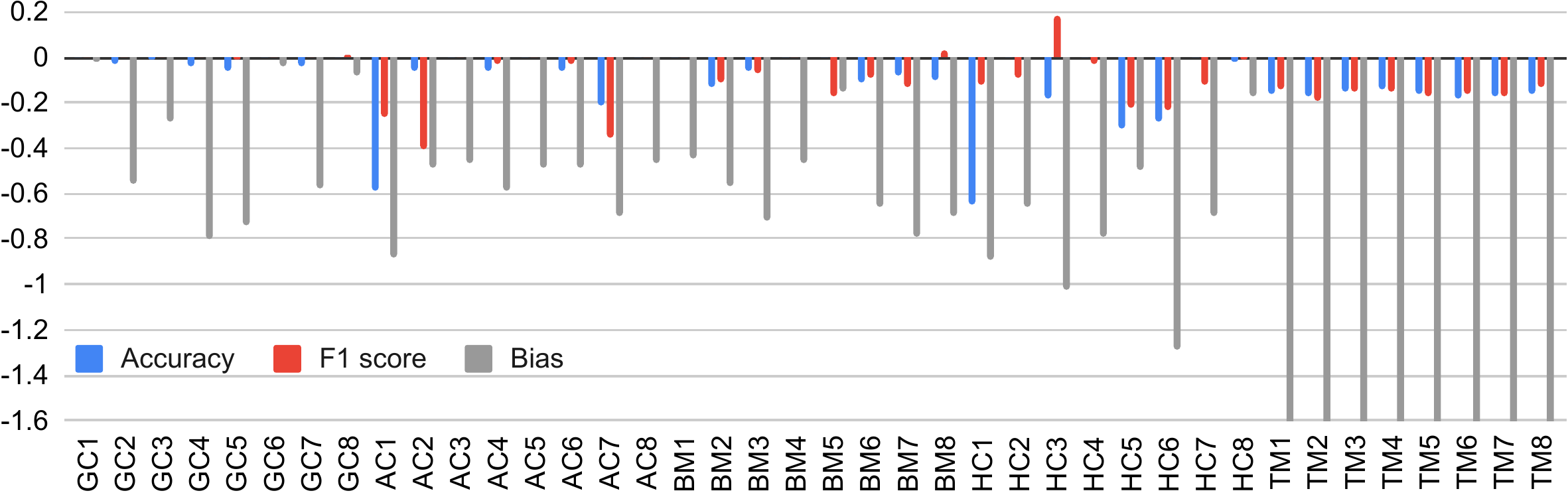}
	\caption{Change of performance and bias after applying bias mitigation technique (negative value indicates reduction)}
	\label{impact}
\end{figure}

While mitigating bias, there is a chance that the performance of the model is diminished. The most successful algorithm in debiasing a model does not always give good performance. So, often the developers have to trade-off between fairness and performance. 
In this section, we have investigated the answer to RQ3. What are the impacts when the bias mitigation algorithms are applied to the models? We have analyzed the accuracy and F1 score of the models after applying the mitigation algorithms.
First, for each model, we have analyzed the impacts of the most effective mitigation algorithms in removing bias. In \fignref{impact}, we have plotted the change in accuracy, F1 score, and total bias when the most successful mitigating algorithms are applied. We can see that while mitigating bias, many models are losing their performance.
From \tabref{tab:main}, pre-processing algorithms, especially Reweighing has been the most effective in model GC1, GC6, AC3, AC5, AC8, BM1, and BM3. From \fignref{impact}, these models always retain their performance after mitigation. \\

\finding{When mitigating bias effectively, in-processing mitigation algorithms show uncertain behavior in their performance.}
Among in-processing algorithms, Adversarial debiasing has been the most effective in 11 models (GC2, GC3, GC4, GC5, AC2, AC7, HC1, HC5, HC6), and Prejudice remover has been the most effective in 1 model (HC2). We have found that for German Credit models Adversarial debiasing has been effective without losing performance. But in other cases, AC1, AC7, HC1, and HC7, the accuracy has decreased at least 21.4\%. 
In HC2, Prejudice remover also loses F1 score while mitigating the bias. 
Since, in-processing techniques employ new model and ignore the prediction of the original model, in all situations (dataset and task), it is not giving better performance. In our evaluation, adversarial debiasing is giving good performance with German Credit dataset but not on Adult Census or Home Credit dataset. Therefore, in-processing techniques are not appropriate when we can not change the original modeling. Also, since these techniques are uncertain in retaining performance, the developers should be careful about the accuracy of prediction after the intervention. \\

\finding{Although post-processing algorithms are the most dominating in debiasing, they are always diminishing the model accuracy and F1 score.}
From \tabref{tab:main}, we can see that in 21 out of 40 models, one of the post-processing algorithms are being the most successful. But in all of the cases they are losing performance. The average accuracy reduction in these models is 7.49\% and average F1 decrease is 10.07\%. For example, in AC1, the most bias mitigating algorithm is Reject option classification but the model is loosing 26.1\% accuracy and 40\% F1 score. In these cases, developers should move to the next best mitigation algorithm.
In a few other cases such as HC8, the Reject Option classification mitigates bias with only 1.6\% loss in accuracy and 1.3\% loss in f1 score. In such situations, post-processing techniques can be applied to mitigate the bias. \\

\begin{figure}[!ht]
	\centering
	\includegraphics[width=\columnwidth]{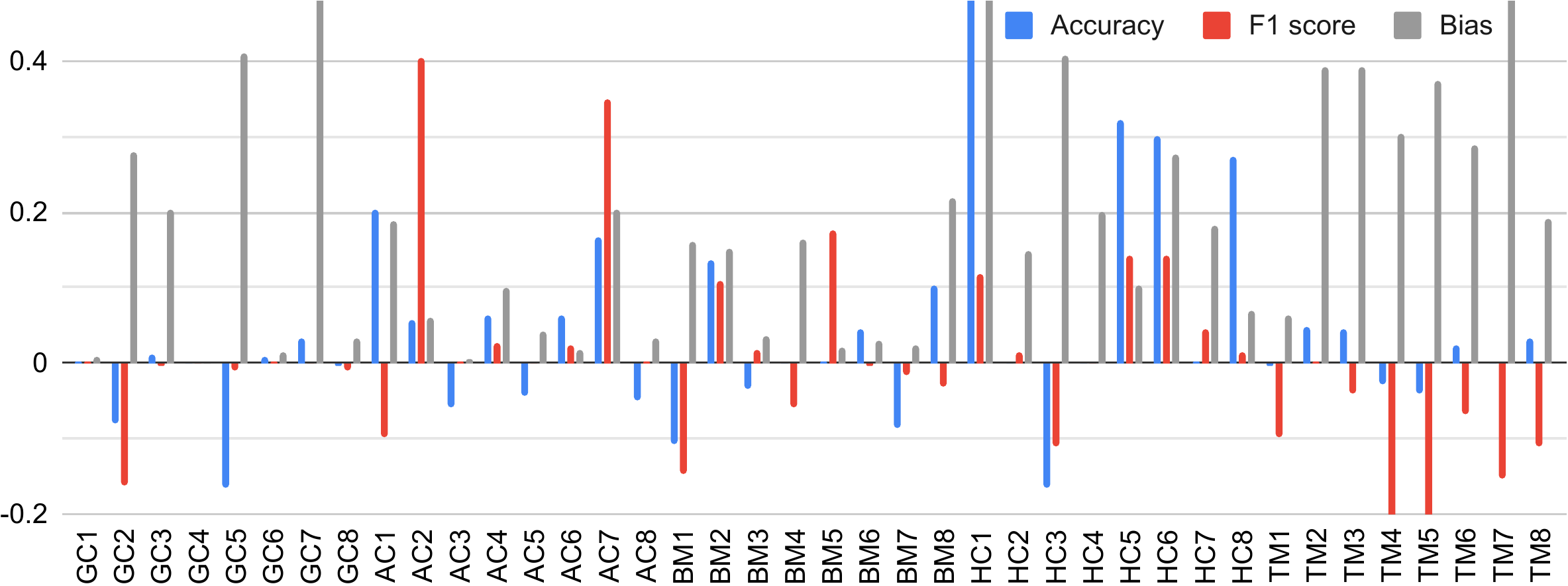}
	\caption{Change of performance and bias between the 1st and 2nd most successful mitigation algorithms (negative value indicates reduction)}
	\label{compare}
\end{figure}

\finding{Trade-off between performance and fairness exists, and post-processing algorithms have most competitive replacement.}
Since some most mitigating algorithms are having performance reduction, for each model, we have compared the most successful algorithm with the next best mitigation algorithm in \fignref{compare}. We have found that for 18 out of 40 models, the performance of the 2nd ranked algorithm is same or better than the 1st ranked algorithm. Among them, in AC4, AC6, BM5, HC5, and HC8, the 2nd ranked algorithm has bias very close (not more than 0.1) to the 1st ranked one. All of these, except HC5, the 1st ranked bias mitigation algorithm is a post-processing technique. We observe that competitive alternative mitigation technique is more common for post-processing mitigation algorithms.
Therefore, if we increase the tolerable range of bias, then other mitigation techniques would be better alternative in terms of performance. 

%% file: threat.tex
\section{Threats to Validity}
\label{sec:threats}

{\em Benchmark Creation.}
To avoid experimenting on low-quality kernels, we have only considered the kernels with more than 5 votes. In addition, we have excluded the kernels where the model accuracy is very low (less than 65\%). Finally, we have selected the top voted ones from the list. We have also verified that the collected kernels are runnable.
To ensure the models collected from Kaggle are appropriate for fairness study, we have first selected the fairness analysis datasets from previous works and searched models for those datasets. Finally, we have searched competitions that use dataset with protected attributes used in the literature. 

{\em Fairness and performance evaluation.}
Our collected models give the same performance, as mentioned in the corresponding Kaggle kernels. For evaluating fairness and applying mitigation algorithms we have used AIF 360 toolkit \cite{bellamy2018ai} developed by IBM. Bellamy \etal presented fairness results (4 metrics) for two models (Logistic regression and Random forest) on Adult Census dataset with protected attribute \textit{race} \cite{bellamy2018ai}. We have done experiment with the same setup and validated our result \cite{bellamy2018ai}. Similar to \cite{friedler2019comparative}, for each metric, we have evaluated 10 times and taken the mean of the values. The stability comparison of the results is shown in \secref{sec:unfairness}.

{\em Fairness comparison.}
As different metrics are computed based on different definitions of fairness, we have compared bias using a specific metric or cumulatively. Finally, in this paper, we have focused on comparing fairness of different models. Therefore, for each dataset, we followed the same method to pre-process training and testing data. 

%% file: related.tex
\section{Related Works}
\label{sec:related}

{\em SE for Fairness in ML.}
This line of work is the closest to our work. 
FairTest~\cite{tramer2017fairtest} proposes methodology to detect unwarranted feature associations and potential biases in a dataset using manually written tests.  
Themis~\cite{galhotra2017fairness} generates random tests automatically to detect causal fairness using black-box decision making process. 
Aequitas~\cite{udeshi2018automated} is a fully automated directed test generation module to generate discriminatory inputs in ML models, which can be used to validate individual fairness. 
FairML~\cite{adebayo2016iterative} introduces an orthogonal transformation methodology to quantify the relative dependence of black-box models to its input features, with the goal of assessing fairness.
A more recent work \cite{aggarwal2019black} proposes black-box fairness testing method to detect individual discrimination in ML models. They \cite{aggarwal2019black} propose a test case generation algorithm based on symbolic execution and local explainability.
The above works have proposed novel techniques to detect and test fairness in ML systems. However, we have focused on empirical evaluation of fairness in ML models written by practitioners and reported our findings.
Friedler \etal also worked on an empirical study but compared between fairness enhancing interventions and not models~\cite{friedler2019comparative}. Harrison \etal conducted survey based empirical study to understand how fairness of different models is perceived by humans~\cite{harrison2020empirical}. Holstein \etal also conducted survey on industry developers to find the challenges for developing fairness-aware tools and models~\cite{holstein2019improving}. However, no empirical study has been conducted to measure and compare fairness of ML models in practice, and analyze the impacts of mitigation algorithms on the models. 

{\em Fairness measure and algorithms.} 
The machine learning community has focused on novel techniques to identify, measure and mitigate bias \cite{dixon2018measuring, pearl2009causal, dwork2012fairness, feldman2015certifying, calders2010three, zafar2015fairness, hardt2016equality, chouldechova2017fair, goh2016satisfying, kamishima2012fairness}. 
This body of work concentrate on the theoretical aspects of bias in ML classifiers. 
Different fairness measures and mitigation algorithms have been discussed in \secref{subsec:measures} and \secref{subsec:bias-mitigation}.  
In this work, we have focused on the software engineering 
aspects of ML models used in practice.

{\em ML model testing.} 
DeepCheck \cite{gopinath2018symbolic} proposes lightweight white-box symbolic analysis to validate deep neural networks (DNN). 
DeepXplore \cite{pei2017deepxplore} proposes a white-box framework to generate test input that can exploit the incorrect behavior of DNNs. DeepTest \cite{tian2018deeptest} uses domain-specific metamorphic relations to detect errors in 
DNN based software. 
These works have focused on the robustness property of ML systems, whereas we have studied fairness property that is fundamentally different from robustness~\cite{udeshi2018automated}.

%% file: conclusion.tex
\section{Conclusion}
\label{sec:conc}
ML fairness has received much attention recently. However, ML libraries do not provide enough support to address the issue in practice.  
In this paper, we have empirically evaluated the fairness of ML models and discussed our findings of software engineering aspects. First, we have created a benchmark of 40 ML models from 5 different problem domains. Then, we have used a comprehensive set of fairness metrics to measure fairness. After that, we have applied 7 mitigation techniques on the models and computed the fairness metric again. We have also evaluated the performance impact of the models after mitigation techniques are applied. We have found what kind of bias is more common and how they could be addressed. Our study also suggests that further SE research and library enhancements are needed to make fairness concerns more accessible to developers.

%% file: paper.bbl

\begin{thebibliography}{52}


\ifx \showCODEN    \undefined \def \showCODEN     #1{\unskip}     \fi
\ifx \showDOI      \undefined \def \showDOI       #1{#1}\fi
\ifx \showISBNx    \undefined \def \showISBNx     #1{\unskip}     \fi
\ifx \showISBNxiii \undefined \def \showISBNxiii  #1{\unskip}     \fi
\ifx \showISSN     \undefined \def \showISSN      #1{\unskip}     \fi
\ifx \showLCCN     \undefined \def \showLCCN      #1{\unskip}     \fi
\ifx \shownote     \undefined \def \shownote      #1{#1}          \fi
\ifx \showarticletitle \undefined \def \showarticletitle #1{#1}   \fi
\ifx \showURL      \undefined \def \showURL       {\relax}        \fi
\providecommand\bibfield[2]{#2}
\providecommand\bibinfo[2]{#2}
\providecommand\natexlab[1]{#1}
\providecommand\showeprint[2][]{arXiv:#2}

\bibitem[\protect\citeauthoryear{Adebayo and Kagal}{Adebayo and Kagal}{2016}]%
        {adebayo2016iterative}
\bibfield{author}{\bibinfo{person}{Julius Adebayo} {and}
  \bibinfo{person}{Lalana Kagal}.} \bibinfo{year}{2016}\natexlab{}.
\newblock \showarticletitle{Iterative orthogonal feature projection for
  diagnosing bias in black-box models}.
\newblock \bibinfo{journal}{\emph{arXiv preprint arXiv:1611.04967}}
  (\bibinfo{year}{2016}).
\newblock


\bibitem[\protect\citeauthoryear{Adebayo et~al\mbox{.}}{Adebayo
  et~al\mbox{.}}{2016}]%
        {adebayo2016fairml}
\bibfield{author}{\bibinfo{person}{Julius~A Adebayo} {et~al\mbox{.}}}
  \bibinfo{year}{2016}\natexlab{}.
\newblock \emph{\bibinfo{title}{FairML: ToolBox for diagnosing bias in
  predictive modeling}}.
\newblock \bibinfo{thesistype}{Ph.D. Dissertation}.
  \bibinfo{school}{Massachusetts Institute of Technology}.
\newblock


\bibitem[\protect\citeauthoryear{Aggarwal, Lohia, Nagar, Dey, and
  Saha}{Aggarwal et~al\mbox{.}}{2019}]%
        {aggarwal2019black}
\bibfield{author}{\bibinfo{person}{Aniya Aggarwal}, \bibinfo{person}{Pranay
  Lohia}, \bibinfo{person}{Seema Nagar}, \bibinfo{person}{Kuntal Dey}, {and}
  \bibinfo{person}{Diptikalyan Saha}.} \bibinfo{year}{2019}\natexlab{}.
\newblock \showarticletitle{Black box fairness testing of machine learning
  models}. In \bibinfo{booktitle}{\emph{Proceedings of the 2019 27th ACM Joint
  Meeting on European Software Engineering Conference and Symposium on the
  Foundations of Software Engineering}}. \bibinfo{pages}{625--635}.
\newblock


\bibitem[\protect\citeauthoryear{Bellamy, Dey, Hind, Hoffman, Houde, Kannan,
  Lohia, Martino, Mehta, Mojsilovic, et~al\mbox{.}}{Bellamy
  et~al\mbox{.}}{2018}]%
        {bellamy2018ai}
\bibfield{author}{\bibinfo{person}{Rachel~KE Bellamy}, \bibinfo{person}{Kuntal
  Dey}, \bibinfo{person}{Michael Hind}, \bibinfo{person}{Samuel~C Hoffman},
  \bibinfo{person}{Stephanie Houde}, \bibinfo{person}{Kalapriya Kannan},
  \bibinfo{person}{Pranay Lohia}, \bibinfo{person}{Jacquelyn Martino},
  \bibinfo{person}{Sameep Mehta}, \bibinfo{person}{Aleksandra Mojsilovic},
  {et~al\mbox{.}}} \bibinfo{year}{2018}\natexlab{}.
\newblock \showarticletitle{AI Fairness 360: An extensible toolkit for
  detecting, understanding, and mitigating unwanted algorithmic bias}.
\newblock \bibinfo{journal}{\emph{arXiv preprint arXiv:1810.01943}}
  (\bibinfo{year}{2018}).
\newblock


\bibitem[\protect\citeauthoryear{Berk, Heidari, Jabbari, Kearns, and Roth}{Berk
  et~al\mbox{.}}{2018}]%
        {berk2018fairness}
\bibfield{author}{\bibinfo{person}{Richard Berk}, \bibinfo{person}{Hoda
  Heidari}, \bibinfo{person}{Shahin Jabbari}, \bibinfo{person}{Michael Kearns},
  {and} \bibinfo{person}{Aaron Roth}.} \bibinfo{year}{2018}\natexlab{}.
\newblock \showarticletitle{Fairness in criminal justice risk assessments: The
  state of the art}.
\newblock \bibinfo{journal}{\emph{Sociological Methods \& Research}}
  (\bibinfo{year}{2018}), \bibinfo{pages}{0049124118782533}.
\newblock


\bibitem[\protect\citeauthoryear{Binns}{Binns}{2017}]%
        {binns2017fairness}
\bibfield{author}{\bibinfo{person}{Reuben Binns}.}
  \bibinfo{year}{2017}\natexlab{}.
\newblock \showarticletitle{Fairness in machine learning: Lessons from
  political philosophy}.
\newblock \bibinfo{journal}{\emph{arXiv preprint arXiv:1712.03586}}
  (\bibinfo{year}{2017}).
\newblock


\bibitem[\protect\citeauthoryear{Biswas and Rajan}{Biswas and Rajan}{2020}]%
        {biswas20203912064}
\bibfield{author}{\bibinfo{person}{Sumon Biswas} {and} \bibinfo{person}{Hridesh
  Rajan}.} \bibinfo{year}{2020}\natexlab{}.
\newblock \bibinfo{booktitle}{\emph{{ML-Fairness: Accepted Artifact for
  ESEC/FSE 2020 Paper on Fairness of Machine Learning Models}}}.
\newblock
\urldef\tempurl%
\url{https://doi.org/10.5281/zenodo.3912064}
\showDOI{\tempurl}


\bibitem[\protect\citeauthoryear{Calders and Verwer}{Calders and
  Verwer}{2010}]%
        {calders2010three}
\bibfield{author}{\bibinfo{person}{Toon Calders} {and} \bibinfo{person}{Sicco
  Verwer}.} \bibinfo{year}{2010}\natexlab{}.
\newblock \showarticletitle{Three naive Bayes approaches for
  discrimination-free classification}.
\newblock \bibinfo{journal}{\emph{Data Mining and Knowledge Discovery}}
  \bibinfo{volume}{21}, \bibinfo{number}{2} (\bibinfo{year}{2010}),
  \bibinfo{pages}{277--292}.
\newblock


\bibitem[\protect\citeauthoryear{Carpenter}{Carpenter}{2011}]%
        {carpenter2011may}
\bibfield{author}{\bibinfo{person}{Jennifer Carpenter}.}
  \bibinfo{year}{2011}\natexlab{}.
\newblock \bibinfo{title}{May the best analyst win}.
\newblock \bibinfo{howpublished}{American Association for the Advancement of
  Science}.
\newblock


\bibitem[\protect\citeauthoryear{Chen, Kallus, Mao, Svacha, and Udell}{Chen
  et~al\mbox{.}}{2019}]%
        {chen2019fairness}
\bibfield{author}{\bibinfo{person}{Jiahao Chen}, \bibinfo{person}{Nathan
  Kallus}, \bibinfo{person}{Xiaojie Mao}, \bibinfo{person}{Geoffry Svacha},
  {and} \bibinfo{person}{Madeleine Udell}.} \bibinfo{year}{2019}\natexlab{}.
\newblock \showarticletitle{Fairness under unawareness: Assessing disparity
  when protected class is unobserved}. In \bibinfo{booktitle}{\emph{Proceedings
  of the Conference on Fairness, Accountability, and Transparency}}.
  \bibinfo{pages}{339--348}.
\newblock


\bibitem[\protect\citeauthoryear{Chouldechova}{Chouldechova}{2017}]%
        {chouldechova2017fair}
\bibfield{author}{\bibinfo{person}{Alexandra Chouldechova}.}
  \bibinfo{year}{2017}\natexlab{}.
\newblock \showarticletitle{Fair prediction with disparate impact: A study of
  bias in recidivism prediction instruments}.
\newblock \bibinfo{journal}{\emph{Big data}} \bibinfo{volume}{5},
  \bibinfo{number}{2} (\bibinfo{year}{2017}), \bibinfo{pages}{153--163}.
\newblock


\bibitem[\protect\citeauthoryear{Dahl, Jaitly, and Salakhutdinov}{Dahl
  et~al\mbox{.}}{2014}]%
        {dahl2014multi}
\bibfield{author}{\bibinfo{person}{George~E Dahl}, \bibinfo{person}{Navdeep
  Jaitly}, {and} \bibinfo{person}{Ruslan Salakhutdinov}.}
  \bibinfo{year}{2014}\natexlab{}.
\newblock \showarticletitle{Multi-task neural networks for QSAR predictions}.
\newblock \bibinfo{journal}{\emph{arXiv preprint arXiv:1406.1231}}
  (\bibinfo{year}{2014}).
\newblock


\bibitem[\protect\citeauthoryear{Dixon, Li, Sorensen, Thain, and
  Vasserman}{Dixon et~al\mbox{.}}{2018}]%
        {dixon2018measuring}
\bibfield{author}{\bibinfo{person}{Lucas Dixon}, \bibinfo{person}{John Li},
  \bibinfo{person}{Jeffrey Sorensen}, \bibinfo{person}{Nithum Thain}, {and}
  \bibinfo{person}{Lucy Vasserman}.} \bibinfo{year}{2018}\natexlab{}.
\newblock \showarticletitle{Measuring and mitigating unintended bias in text
  classification}. In \bibinfo{booktitle}{\emph{Proceedings of the 2018
  AAAI/ACM Conference on AI, Ethics, and Society}}. \bibinfo{pages}{67--73}.
\newblock


\bibitem[\protect\citeauthoryear{Dwork, Hardt, Pitassi, Reingold, and
  Zemel}{Dwork et~al\mbox{.}}{2012}]%
        {dwork2012fairness}
\bibfield{author}{\bibinfo{person}{Cynthia Dwork}, \bibinfo{person}{Moritz
  Hardt}, \bibinfo{person}{Toniann Pitassi}, \bibinfo{person}{Omer Reingold},
  {and} \bibinfo{person}{Richard Zemel}.} \bibinfo{year}{2012}\natexlab{}.
\newblock \showarticletitle{Fairness through awareness}. In
  \bibinfo{booktitle}{\emph{Proceedings of the 3rd innovations in theoretical
  computer science conference}}. \bibinfo{pages}{214--226}.
\newblock


\bibitem[\protect\citeauthoryear{Feldman, Friedler, Moeller, Scheidegger, and
  Venkatasubramanian}{Feldman et~al\mbox{.}}{2015}]%
        {feldman2015certifying}
\bibfield{author}{\bibinfo{person}{Michael Feldman}, \bibinfo{person}{Sorelle~A
  Friedler}, \bibinfo{person}{John Moeller}, \bibinfo{person}{Carlos
  Scheidegger}, {and} \bibinfo{person}{Suresh Venkatasubramanian}.}
  \bibinfo{year}{2015}\natexlab{}.
\newblock \showarticletitle{Certifying and removing disparate impact}. In
  \bibinfo{booktitle}{\emph{proceedings of the 21th ACM SIGKDD international
  conference on knowledge discovery and data mining}}.
  \bibinfo{pages}{259--268}.
\newblock


\bibitem[\protect\citeauthoryear{Friedler, Scheidegger, Venkatasubramanian,
  Choudhary, Hamilton, and Roth}{Friedler et~al\mbox{.}}{2019}]%
        {friedler2019comparative}
\bibfield{author}{\bibinfo{person}{Sorelle~A Friedler}, \bibinfo{person}{Carlos
  Scheidegger}, \bibinfo{person}{Suresh Venkatasubramanian},
  \bibinfo{person}{Sonam Choudhary}, \bibinfo{person}{Evan~P Hamilton}, {and}
  \bibinfo{person}{Derek Roth}.} \bibinfo{year}{2019}\natexlab{}.
\newblock \showarticletitle{A comparative study of fairness-enhancing
  interventions in machine learning}. In \bibinfo{booktitle}{\emph{Proceedings
  of the Conference on Fairness, Accountability, and Transparency}}.
  \bibinfo{pages}{329--338}.
\newblock


\bibitem[\protect\citeauthoryear{Galhotra, Brun, and Meliou}{Galhotra
  et~al\mbox{.}}{2017}]%
        {galhotra2017fairness}
\bibfield{author}{\bibinfo{person}{Sainyam Galhotra}, \bibinfo{person}{Yuriy
  Brun}, {and} \bibinfo{person}{Alexandra Meliou}.}
  \bibinfo{year}{2017}\natexlab{}.
\newblock \showarticletitle{Fairness testing: testing software for
  discrimination}. In \bibinfo{booktitle}{\emph{Proceedings of the 2017 11th
  Joint Meeting on Foundations of Software Engineering}}.
  \bibinfo{pages}{498--510}.
\newblock


\bibitem[\protect\citeauthoryear{Goh, Cotter, Gupta, and Friedlander}{Goh
  et~al\mbox{.}}{2016}]%
        {goh2016satisfying}
\bibfield{author}{\bibinfo{person}{Gabriel Goh}, \bibinfo{person}{Andrew
  Cotter}, \bibinfo{person}{Maya Gupta}, {and} \bibinfo{person}{Michael~P
  Friedlander}.} \bibinfo{year}{2016}\natexlab{}.
\newblock \showarticletitle{Satisfying real-world goals with dataset
  constraints}. In \bibinfo{booktitle}{\emph{Advances in Neural Information
  Processing Systems}}. \bibinfo{pages}{2415--2423}.
\newblock


\bibitem[\protect\citeauthoryear{Gopinath, Wang, Zhang, Pasareanu, and
  Khurshid}{Gopinath et~al\mbox{.}}{2018}]%
        {gopinath2018symbolic}
\bibfield{author}{\bibinfo{person}{Divya Gopinath}, \bibinfo{person}{Kaiyuan
  Wang}, \bibinfo{person}{Mengshi Zhang}, \bibinfo{person}{Corina~S Pasareanu},
  {and} \bibinfo{person}{Sarfraz Khurshid}.} \bibinfo{year}{2018}\natexlab{}.
\newblock \showarticletitle{Symbolic execution for deep neural networks}.
\newblock \bibinfo{journal}{\emph{arXiv preprint arXiv:1807.10439}}
  (\bibinfo{year}{2018}).
\newblock


\bibitem[\protect\citeauthoryear{Hardt, Price, and Srebro}{Hardt
  et~al\mbox{.}}{2016}]%
        {hardt2016equality}
\bibfield{author}{\bibinfo{person}{Moritz Hardt}, \bibinfo{person}{Eric Price},
  {and} \bibinfo{person}{Nati Srebro}.} \bibinfo{year}{2016}\natexlab{}.
\newblock \showarticletitle{Equality of opportunity in supervised learning}. In
  \bibinfo{booktitle}{\emph{Advances in neural information processing
  systems}}. \bibinfo{pages}{3315--3323}.
\newblock


\bibitem[\protect\citeauthoryear{Harrison, Hanson, Jacinto, Ramirez, and
  Ur}{Harrison et~al\mbox{.}}{2020}]%
        {harrison2020empirical}
\bibfield{author}{\bibinfo{person}{Galen Harrison}, \bibinfo{person}{Julia
  Hanson}, \bibinfo{person}{Christine Jacinto}, \bibinfo{person}{Julio
  Ramirez}, {and} \bibinfo{person}{Blase Ur}.} \bibinfo{year}{2020}\natexlab{}.
\newblock \showarticletitle{An empirical study on the perceived fairness of
  realistic, imperfect machine learning models}. In
  \bibinfo{booktitle}{\emph{Proceedings of the 2020 Conference on Fairness,
  Accountability, and Transparency}}. \bibinfo{pages}{392--402}.
\newblock


\bibitem[\protect\citeauthoryear{Holstein, Wortman~Vaughan, Daum{\'e}~III,
  Dudik, and Wallach}{Holstein et~al\mbox{.}}{2019}]%
        {holstein2019improving}
\bibfield{author}{\bibinfo{person}{Kenneth Holstein}, \bibinfo{person}{Jennifer
  Wortman~Vaughan}, \bibinfo{person}{Hal Daum{\'e}~III}, \bibinfo{person}{Miro
  Dudik}, {and} \bibinfo{person}{Hanna Wallach}.}
  \bibinfo{year}{2019}\natexlab{}.
\newblock \showarticletitle{Improving fairness in machine learning systems:
  What do industry practitioners need?}. In
  \bibinfo{booktitle}{\emph{Proceedings of the 2019 CHI Conference on Human
  Factors in Computing Systems}}. \bibinfo{pages}{1--16}.
\newblock


\bibitem[\protect\citeauthoryear{Jepsen}{Jepsen}{2014}]%
        {higs-boson}
\bibfield{author}{\bibinfo{person}{Kathryn Jepsen}.}
  \bibinfo{year}{2014}\natexlab{}.
\newblock \bibinfo{title}{The machine learning community takes on the Higgs}.
\newblock
  \bibinfo{howpublished}{\url{https://www.symmetrymagazine.org/article/july-2014/the-machine-learning-community-takes-on-the-higgs}}.
\newblock


\bibitem[\protect\citeauthoryear{Joshi}{Joshi}{2017}]%
        {joshi2017artificial}
\bibfield{author}{\bibinfo{person}{Prateek Joshi}.}
  \bibinfo{year}{2017}\natexlab{}.
\newblock \bibinfo{booktitle}{\emph{Artificial intelligence with python}}.
\newblock \bibinfo{publisher}{Packt Publishing Ltd}.
\newblock


\bibitem[\protect\citeauthoryear{{Kaggle}}{{Kaggle}}{2010}]%
        {kaggle}
\bibfield{author}{\bibinfo{person}{{Kaggle}}.} \bibinfo{year}{2010}\natexlab{}.
\newblock \bibinfo{title}{{The world's largest data science community with
  powerful tools and resources to help you achieve your data science goals.}}
\newblock
\newblock
\newblock
\shownote{\url{www.kaggle.com}.}


\bibitem[\protect\citeauthoryear{{Kaggle}}{{Kaggle}}{2017a}]%
        {adult}
\bibfield{author}{\bibinfo{person}{{Kaggle}}.}
  \bibinfo{year}{2017}\natexlab{a}.
\newblock \bibinfo{title}{{Adult Census Dataset}}.
\newblock
\newblock
\newblock
\shownote{\url{https://www.kaggle.com/uciml/adult-census-income}.}


\bibitem[\protect\citeauthoryear{{Kaggle}}{{Kaggle}}{2017b}]%
        {bank}
\bibfield{author}{\bibinfo{person}{{Kaggle}}.}
  \bibinfo{year}{2017}\natexlab{b}.
\newblock \bibinfo{title}{{Bank Marketing Dataset}}.
\newblock
\newblock
\newblock
\shownote{\url{https://www.kaggle.com/c/bank-marketing-uci}.}


\bibitem[\protect\citeauthoryear{{Kaggle}}{{Kaggle}}{2017c}]%
        {cspr}
\bibfield{author}{\bibinfo{person}{{Kaggle}}.}
  \bibinfo{year}{2017}\natexlab{c}.
\newblock \bibinfo{title}{{Competition: Santander Product Recommendation}}.
\newblock
\newblock
\newblock
\shownote{\url{https://www.kaggle.com/c/santander-product-recommendation/overview}.}


\bibitem[\protect\citeauthoryear{{Kaggle}}{{Kaggle}}{2017d}]%
        {gc}
\bibfield{author}{\bibinfo{person}{{Kaggle}}.}
  \bibinfo{year}{2017}\natexlab{d}.
\newblock \bibinfo{title}{{German Credit Dataset}}.
\newblock
\newblock
\newblock
\shownote{\url{https://www.kaggle.com/uciml/german-credit}.}


\bibitem[\protect\citeauthoryear{{Kaggle}}{{Kaggle}}{2017e}]%
        {home}
\bibfield{author}{\bibinfo{person}{{Kaggle}}.}
  \bibinfo{year}{2017}\natexlab{e}.
\newblock \bibinfo{title}{{Home Credit Dataset}}.
\newblock
\newblock
\newblock
\shownote{\url{https://www.kaggle.com/c/home-credit-default-risk}.}


\bibitem[\protect\citeauthoryear{{Kaggle}}{{Kaggle}}{2017f}]%
        {titanic}
\bibfield{author}{\bibinfo{person}{{Kaggle}}.}
  \bibinfo{year}{2017}\natexlab{f}.
\newblock \bibinfo{title}{{Titanic ML Dataset}}.
\newblock
\newblock
\newblock
\shownote{\url{https://www.kaggle.com/c/titanic/data}.}


\bibitem[\protect\citeauthoryear{{Kaggle}}{{Kaggle}}{2019a}]%
        {kernel-mmta}
\bibfield{author}{\bibinfo{person}{{Kaggle}}.}
  \bibinfo{year}{2019}\natexlab{a}.
\newblock \bibinfo{title}{{Adult Census Kernel: Multiple ML Techniques and
  Analysis}}.
\newblock
\newblock
\newblock
\shownote{\url{https://www.kaggle.com/bananuhbeatdown/multiple-ml-techniques-and-analysis-of-dataset}.}


\bibitem[\protect\citeauthoryear{{Kaggle}}{{Kaggle}}{2019b}]%
        {kernel-gcra}
\bibfield{author}{\bibinfo{person}{{Kaggle}}.}
  \bibinfo{year}{2019}\natexlab{b}.
\newblock \bibinfo{title}{{Kernel: German Credit Risk Analysis}}.
\newblock
\newblock
\newblock
\shownote{\url{https://www.kaggle.com/pahulpreet/german-credit-risk-analysis-beginner-s-guide}.}


\bibitem[\protect\citeauthoryear{Kamiran and Calders}{Kamiran and
  Calders}{2012}]%
        {kamiran2012data}
\bibfield{author}{\bibinfo{person}{Faisal Kamiran} {and} \bibinfo{person}{Toon
  Calders}.} \bibinfo{year}{2012}\natexlab{}.
\newblock \showarticletitle{Data preprocessing techniques for classification
  without discrimination}.
\newblock \bibinfo{journal}{\emph{Knowledge and Information Systems}}
  \bibinfo{volume}{33}, \bibinfo{number}{1} (\bibinfo{year}{2012}),
  \bibinfo{pages}{1--33}.
\newblock


\bibitem[\protect\citeauthoryear{Kamiran, Karim, and Zhang}{Kamiran
  et~al\mbox{.}}{2012}]%
        {kamiran2012decision}
\bibfield{author}{\bibinfo{person}{Faisal Kamiran}, \bibinfo{person}{Asim
  Karim}, {and} \bibinfo{person}{Xiangliang Zhang}.}
  \bibinfo{year}{2012}\natexlab{}.
\newblock \showarticletitle{Decision theory for discrimination-aware
  classification}. In \bibinfo{booktitle}{\emph{2012 IEEE 12th International
  Conference on Data Mining}}. IEEE, \bibinfo{pages}{924--929}.
\newblock


\bibitem[\protect\citeauthoryear{Kamishima, Akaho, Asoh, and Sakuma}{Kamishima
  et~al\mbox{.}}{2012}]%
        {kamishima2012fairness}
\bibfield{author}{\bibinfo{person}{Toshihiro Kamishima},
  \bibinfo{person}{Shotaro Akaho}, \bibinfo{person}{Hideki Asoh}, {and}
  \bibinfo{person}{Jun Sakuma}.} \bibinfo{year}{2012}\natexlab{}.
\newblock \showarticletitle{Fairness-aware classifier with prejudice remover
  regularizer}. In \bibinfo{booktitle}{\emph{Joint European Conference on
  Machine Learning and Knowledge Discovery in Databases}}. Springer,
  \bibinfo{pages}{35--50}.
\newblock


\bibitem[\protect\citeauthoryear{Kleinberg, Mullainathan, and
  Raghavan}{Kleinberg et~al\mbox{.}}{2016}]%
        {kleinberg2016inherent}
\bibfield{author}{\bibinfo{person}{Jon Kleinberg}, \bibinfo{person}{Sendhil
  Mullainathan}, {and} \bibinfo{person}{Manish Raghavan}.}
  \bibinfo{year}{2016}\natexlab{}.
\newblock \showarticletitle{Inherent trade-offs in the determination of risk
  scores}.
\newblock \bibinfo{journal}{\emph{arXiv preprint arXiv:1609.05807}}
  (\bibinfo{year}{2016}).
\newblock


\bibitem[\protect\citeauthoryear{Pearl et~al\mbox{.}}{Pearl
  et~al\mbox{.}}{2009}]%
        {pearl2009causal}
\bibfield{author}{\bibinfo{person}{Judea Pearl} {et~al\mbox{.}}}
  \bibinfo{year}{2009}\natexlab{}.
\newblock \showarticletitle{Causal inference in statistics: An overview}.
\newblock \bibinfo{journal}{\emph{Statistics surveys}}  \bibinfo{volume}{3}
  (\bibinfo{year}{2009}), \bibinfo{pages}{96--146}.
\newblock


\bibitem[\protect\citeauthoryear{Pedreshi, Ruggieri, and Turini}{Pedreshi
  et~al\mbox{.}}{2008}]%
        {pedreshi2008discrimination}
\bibfield{author}{\bibinfo{person}{Dino Pedreshi}, \bibinfo{person}{Salvatore
  Ruggieri}, {and} \bibinfo{person}{Franco Turini}.}
  \bibinfo{year}{2008}\natexlab{}.
\newblock \showarticletitle{Discrimination-aware data mining}. In
  \bibinfo{booktitle}{\emph{Proceedings of the 14th ACM SIGKDD international
  conference on Knowledge discovery and data mining}}.
  \bibinfo{pages}{560--568}.
\newblock


\bibitem[\protect\citeauthoryear{Pei, Cao, Yang, and Jana}{Pei
  et~al\mbox{.}}{2017}]%
        {pei2017deepxplore}
\bibfield{author}{\bibinfo{person}{Kexin Pei}, \bibinfo{person}{Yinzhi Cao},
  \bibinfo{person}{Junfeng Yang}, {and} \bibinfo{person}{Suman Jana}.}
  \bibinfo{year}{2017}\natexlab{}.
\newblock \showarticletitle{Deepxplore: Automated whitebox testing of deep
  learning systems}. In \bibinfo{booktitle}{\emph{proceedings of the 26th
  Symposium on Operating Systems Principles}}. \bibinfo{pages}{1--18}.
\newblock


\bibitem[\protect\citeauthoryear{Pleiss, Raghavan, Wu, Kleinberg, and
  Weinberger}{Pleiss et~al\mbox{.}}{2017}]%
        {pleiss2017fairness}
\bibfield{author}{\bibinfo{person}{Geoff Pleiss}, \bibinfo{person}{Manish
  Raghavan}, \bibinfo{person}{Felix Wu}, \bibinfo{person}{Jon Kleinberg}, {and}
  \bibinfo{person}{Kilian~Q Weinberger}.} \bibinfo{year}{2017}\natexlab{}.
\newblock \showarticletitle{On fairness and calibration}. In
  \bibinfo{booktitle}{\emph{Advances in Neural Information Processing
  Systems}}. \bibinfo{pages}{5680--5689}.
\newblock


\bibitem[\protect\citeauthoryear{{Scikit Learn}}{{Scikit Learn}}{2019a}]%
        {lgbm}
\bibfield{author}{\bibinfo{person}{{Scikit Learn}}.}
  \bibinfo{year}{2019}\natexlab{a}.
\newblock \bibinfo{title}{{LightGBM API Documentation}}.
\newblock
\newblock
\newblock
\shownote{\url{https://lightgbm.readthedocs.io/en/latest/pythonapi/lightgbm.LGBMClassifier.html}.}


\bibitem[\protect\citeauthoryear{{Scikit Learn}}{{Scikit Learn}}{2019b}]%
        {svc}
\bibfield{author}{\bibinfo{person}{{Scikit Learn}}.}
  \bibinfo{year}{2019}\natexlab{b}.
\newblock \bibinfo{title}{{SVC API Documentation}}.
\newblock
\newblock
\newblock
\shownote{\url{https://scikit-learn.org/stable/modules/generated/sklearn.preprocessing.StandardScaler.html}.}


\bibitem[\protect\citeauthoryear{Sokol, Santos-Rodriguez, and Flach}{Sokol
  et~al\mbox{.}}{2019}]%
        {sokol2019fat}
\bibfield{author}{\bibinfo{person}{Kacper Sokol}, \bibinfo{person}{Raul
  Santos-Rodriguez}, {and} \bibinfo{person}{Peter Flach}.}
  \bibinfo{year}{2019}\natexlab{}.
\newblock \showarticletitle{FAT Forensics: A Python Toolbox for Algorithmic
  Fairness, Accountability and Transparency}.
\newblock \bibinfo{journal}{\emph{arXiv preprint arXiv:1909.05167}}
  (\bibinfo{year}{2019}).
\newblock


\bibitem[\protect\citeauthoryear{Speicher, Heidari, Grgic-Hlaca, Gummadi,
  Singla, Weller, and Zafar}{Speicher et~al\mbox{.}}{2018}]%
        {speicher2018unified}
\bibfield{author}{\bibinfo{person}{Till Speicher}, \bibinfo{person}{Hoda
  Heidari}, \bibinfo{person}{Nina Grgic-Hlaca}, \bibinfo{person}{Krishna~P
  Gummadi}, \bibinfo{person}{Adish Singla}, \bibinfo{person}{Adrian Weller},
  {and} \bibinfo{person}{Muhammad~Bilal Zafar}.}
  \bibinfo{year}{2018}\natexlab{}.
\newblock \showarticletitle{A unified approach to quantifying algorithmic
  unfairness: Measuring individual \&group unfairness via inequality indices}.
  In \bibinfo{booktitle}{\emph{Proceedings of the 24th ACM SIGKDD International
  Conference on Knowledge Discovery \& Data Mining}}.
  \bibinfo{pages}{2239--2248}.
\newblock


\bibitem[\protect\citeauthoryear{{Stack Overflow}}{{Stack Overflow}}{2016}]%
        {class-weight}
\bibfield{author}{\bibinfo{person}{{Stack Overflow}}.}
  \bibinfo{year}{2016}\natexlab{}.
\newblock \bibinfo{title}{{How does the class\_weight parameter in scikit-learn
  work?}}
\newblock
\newblock
\newblock
\shownote{\url{https://stackoverflow.com/questions/30972029/how-does-the-class-weight-parameter-in-scikit-learn-work}.}


\bibitem[\protect\citeauthoryear{Tian, Pei, Jana, and Ray}{Tian
  et~al\mbox{.}}{2018}]%
        {tian2018deeptest}
\bibfield{author}{\bibinfo{person}{Yuchi Tian}, \bibinfo{person}{Kexin Pei},
  \bibinfo{person}{Suman Jana}, {and} \bibinfo{person}{Baishakhi Ray}.}
  \bibinfo{year}{2018}\natexlab{}.
\newblock \showarticletitle{Deeptest: Automated testing of
  deep-neural-network-driven autonomous cars}. In
  \bibinfo{booktitle}{\emph{Proceedings of the 40th international conference on
  software engineering}}. \bibinfo{pages}{303--314}.
\newblock


\bibitem[\protect\citeauthoryear{Tramer, Atlidakis, Geambasu, Hsu, Hubaux,
  Humbert, Juels, and Lin}{Tramer et~al\mbox{.}}{2017}]%
        {tramer2017fairtest}
\bibfield{author}{\bibinfo{person}{Florian Tramer}, \bibinfo{person}{Vaggelis
  Atlidakis}, \bibinfo{person}{Roxana Geambasu}, \bibinfo{person}{Daniel Hsu},
  \bibinfo{person}{Jean-Pierre Hubaux}, \bibinfo{person}{Mathias Humbert},
  \bibinfo{person}{Ari Juels}, {and} \bibinfo{person}{Huang Lin}.}
  \bibinfo{year}{2017}\natexlab{}.
\newblock \showarticletitle{FairTest: Discovering unwarranted associations in
  data-driven applications}. In \bibinfo{booktitle}{\emph{2017 IEEE European
  Symposium on Security and Privacy (EuroS\&P)}}. IEEE,
  \bibinfo{pages}{401--416}.
\newblock


\bibitem[\protect\citeauthoryear{Udeshi, Arora, and Chattopadhyay}{Udeshi
  et~al\mbox{.}}{2018}]%
        {udeshi2018automated}
\bibfield{author}{\bibinfo{person}{Sakshi Udeshi}, \bibinfo{person}{Pryanshu
  Arora}, {and} \bibinfo{person}{Sudipta Chattopadhyay}.}
  \bibinfo{year}{2018}\natexlab{}.
\newblock \showarticletitle{Automated directed fairness testing}. In
  \bibinfo{booktitle}{\emph{Proceedings of the 33rd ACM/IEEE International
  Conference on Automated Software Engineering}}. \bibinfo{pages}{98--108}.
\newblock


\bibitem[\protect\citeauthoryear{Zafar, Valera, Rodriguez, and Gummadi}{Zafar
  et~al\mbox{.}}{2015}]%
        {zafar2015fairness}
\bibfield{author}{\bibinfo{person}{Muhammad~Bilal Zafar},
  \bibinfo{person}{Isabel Valera}, \bibinfo{person}{Manuel~Gomez Rodriguez},
  {and} \bibinfo{person}{Krishna~P Gummadi}.} \bibinfo{year}{2015}\natexlab{}.
\newblock \showarticletitle{Fairness constraints: Mechanisms for fair
  classification}.
\newblock \bibinfo{journal}{\emph{arXiv preprint arXiv:1507.05259}}
  (\bibinfo{year}{2015}).
\newblock


\bibitem[\protect\citeauthoryear{Zemel, Wu, Swersky, Pitassi, and Dwork}{Zemel
  et~al\mbox{.}}{2013}]%
        {zemel2013learning}
\bibfield{author}{\bibinfo{person}{Rich Zemel}, \bibinfo{person}{Yu Wu},
  \bibinfo{person}{Kevin Swersky}, \bibinfo{person}{Toni Pitassi}, {and}
  \bibinfo{person}{Cynthia Dwork}.} \bibinfo{year}{2013}\natexlab{}.
\newblock \showarticletitle{Learning fair representations}. In
  \bibinfo{booktitle}{\emph{International Conference on Machine Learning}}.
  \bibinfo{pages}{325--333}.
\newblock


\bibitem[\protect\citeauthoryear{Zhang, Lemoine, and Mitchell}{Zhang
  et~al\mbox{.}}{2018}]%
        {zhang2018mitigating}
\bibfield{author}{\bibinfo{person}{Brian~Hu Zhang}, \bibinfo{person}{Blake
  Lemoine}, {and} \bibinfo{person}{Margaret Mitchell}.}
  \bibinfo{year}{2018}\natexlab{}.
\newblock \showarticletitle{Mitigating unwanted biases with adversarial
  learning}. In \bibinfo{booktitle}{\emph{Proceedings of the 2018 AAAI/ACM
  Conference on AI, Ethics, and Society}}. \bibinfo{pages}{335--340}.
\newblock


\end{thebibliography}
